\documentclass[10pt,twocolumn,letterpaper]{article}

\usepackage[pagenumbers]{cvpr} 

\definecolor{cvprblue}{rgb}{0.21,0.49,0.74}
\usepackage[pagebackref,breaklinks,colorlinks,allcolors=cvprblue]{hyperref}

\usepackage[table]{xcolor}
\usepackage{multirow}
\usepackage{amsmath}
\DeclareMathOperator*{\argmin}{arg\,min}
\usepackage{dsfont}
\usepackage{graphicx}
\usepackage{array}
\usepackage{cuted}
\usepackage{caption}
\usepackage{listings}
\usepackage{textcomp}
\usepackage{siunitx}
\usepackage{tabularx}

\def\confName{CVPR}
\def\confYear{2027}

\title{InstanceRecon: Instance-Decomposed Scene Reconstruction from Sparse Views}

\author{%
  Katharina Schmid\textsuperscript{1} \quad
  Nicolas von Lützow\textsuperscript{1} \quad
  Jozef Hladký\textsuperscript{2} \quad
  Angela Dai\textsuperscript{1} \quad
  Matthias Nie{\ss}ner\textsuperscript{1} \\[0.6em]
  \normalsize
  \textsuperscript{1}Technical University of Munich \quad
  \textsuperscript{2}Computing Systems Lab, Huawei Technologies, Switzerland
}

\begin{document}
\maketitle

\begin{strip}
  \vspace{-3em}
  \centering
  \includegraphics[width=\textwidth]{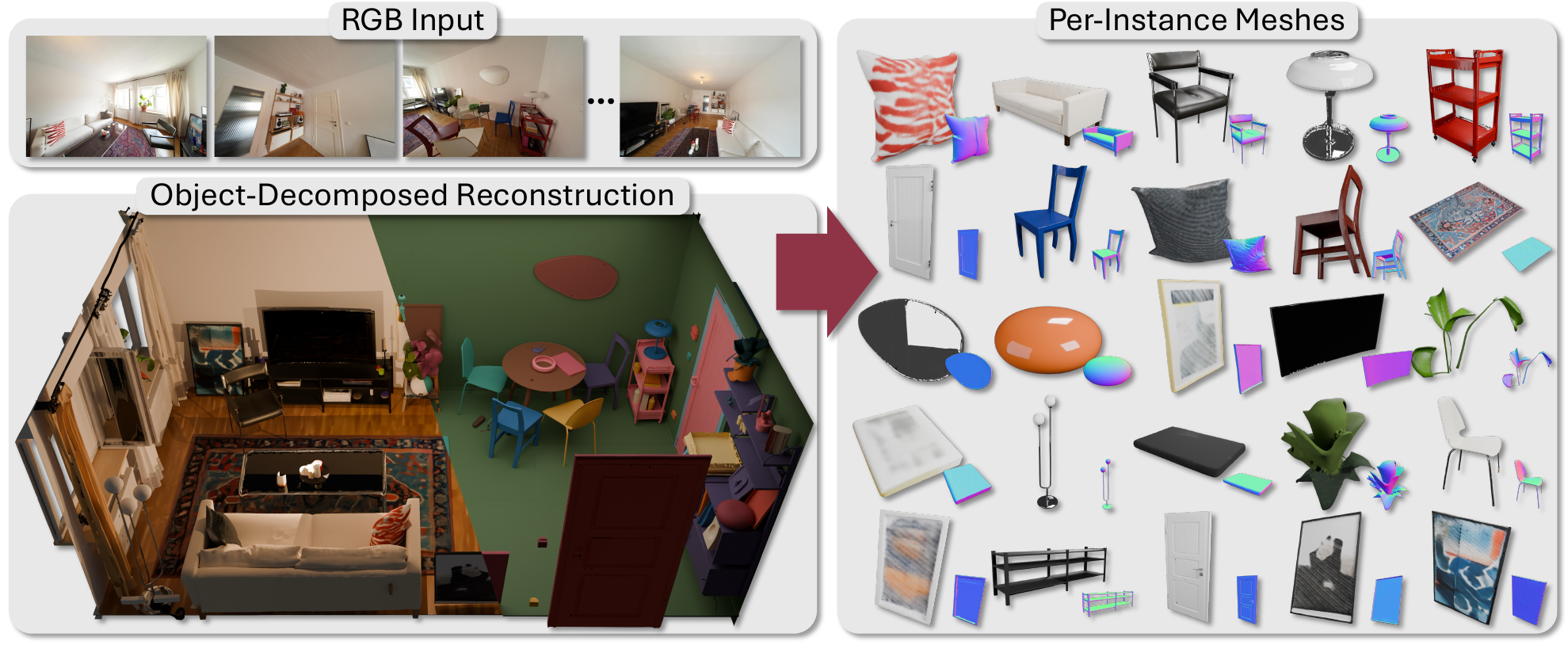}
  \captionsetup{hypcap=false}
  \captionof{figure}{\textbf{InstanceRecon.} Given a sparse set of posed RGB images (top left), our method reconstructs the scene (bottom left) as a set of complete, high-fidelity per-instance PBR meshes (right). All objects are generated jointly by a single scene-level model, so they are consistently placed in the scene, yet each object is a separate, complete mesh, including surfaces that are occluded or hidden where objects touch.}
  \captionsetup{hypcap=true}
  \label{fig:overview}
\end{strip}

\begin{abstract}
We introduce a new approach to instance-decomposed 3D scene reconstruction from sparse multi-view RGB images, producing indoor scenes whose objects are individually complete, editable, and high-fidelity.
Rather than reconstructing objects independently and composing them afterward, we formulate scene reconstruction as the joint generation of N volumes in a shared coordinate frame. 
We distribute instances across volumes via proximity-weighted graph coloring with soft constraints: touching objects are placed in different volumes, and where no valid coloring exists, only the least-close contacts share one. 
Each volume thus contains a variable number of spatially separated objects, keeping $N$ small and independent of scene complexity, while each object can be extracted as a complete 3D entity.
We lift a strong object-level 3D generative prior to this representation and jointly infer object geometry and spatial arrangement with a single scene-level model, guided by per-volume geometry conditioning.
This joint formulation improves reconstruction quality over compositional approaches, while the volume-based representation scales to scenes with an arbitrarily large number of objects.
\end{abstract}    
\section{Introduction}
\label{sec:intro}

We address the task of reconstructing object-decomposed indoor scenes from a sparse set of posed multi-view RGB images. 
Conventional 3D reconstruction recovers the scene as a single surface, which suffices for viewing but not for interaction: objects cannot be moved, edited, or simulated individually.
Our goal is therefore a scene in which every object is a complete, separate 3D entity.
Such digital replicas of real environments are valuable across many domains: a robot could be trained to act in a user's actual living room, an occupant could virtually rearrange furniture to explore alternative layouts, an engineer could simulate how a real house responds to an earthquake, and a designer could turn a captured space into an editable setting for a game or virtual experience.
Today, however, producing them typically requires specialized scanners or extensive manual modeling.
Reconstructing them from a handful of casually captured photographs would make them widely accessible.

Reconstructing such scenes from sparse images, however, is a highly underconstrained inverse problem.
Recovering complete 3D geometry from 2D observations is difficult in itself, and with only a few views, large parts of the scene remain unobserved.
In addition, the scene must be decomposed into individual objects, a task that is ambiguous in cluttered environments with heavy occlusion.
Each object must be reconstructed completely, including surfaces that are never visible, such as contact regions between adjacent objects.
Finally, the inferred structure must remain consistent with the observations and across objects.

Recent approaches increasingly leverage strong object-level 3D generative priors to reconstruct scenes compositionally, generating individual objects and assembling them into a common scene~\cite{siddiquiShapeRRobustConditional2026, xiaSimReconSimReadyCompositional2026, dongReplicateAnySceneZeroShotVideoto3D2026, niuWorldSculptGeneratingCompositional2026}. 
While this yields impressive object-level fidelity, independent reconstruction often introduces inconsistencies in relative position, scale, and orientation, resulting in inaccurate spatial relationships and globally incoherent scenes.
A smaller body of work instead reconstructs all object instances jointly, for example, by jointly optimizing per-object SDFs or jointly generating per-instance latent representations~\cite{niDecompositionalNeuralScene2025, yangInstasceneComplete3d2025, zhuObjectGSObjectawareScene2025, wuObjectSDFImprovedObjectCompositional2023}. 
By modeling interactions between objects during reconstruction, these approaches can improve global scene consistency. 
However, their computational cost typically grows with the number of object instances, limiting them to scenes with relatively few objects and making it difficult to scale to more complex environments.

In this work, we jointly reconstruct all objects in a scene, at a generation cost that does not scale with their number.
Our key observation is that objects only need to be separated where they touch.
We therefore represent a scene by a small, fixed number of $N$ volumes in a shared coordinate frame, each holding a variable number of spatially separated objects.
We assign objects to volumes via proximity-weighted graph coloring, so that touching objects are generated in different volumes and every object, including its contact surfaces, can be extracted as a complete mesh.
We generate all volumes jointly by lifting a state-of-the-art object-level 3D generative prior to this representation, and condition each volume on its own TSDF from predicted depth and instance masks to resolve which volume holds which object.

To summarize, our contributions are:
\begin{itemize}
    \item{We introduce an $N$-volume scene representation that separates touching objects into a small,
      fixed number of volumes via weighted graph coloring with soft constraints. It yields complete,
      separable instance meshes at a generation cost independent of the number of objects.}
    \item{We lift a 3D object generative prior to this decomposed scene setting by denoising all volumes jointly in a single scene-level transformer.}
    \item{We condition each volume on its own TSDF fused from predicted depth and instance masks,
      which resolves the ambiguity of which volume generates which object.}
\end{itemize}

\section{Related Work}
\label{sec:related}
\begin{figure*}[t]
  \centering
  \includegraphics[width=\textwidth]{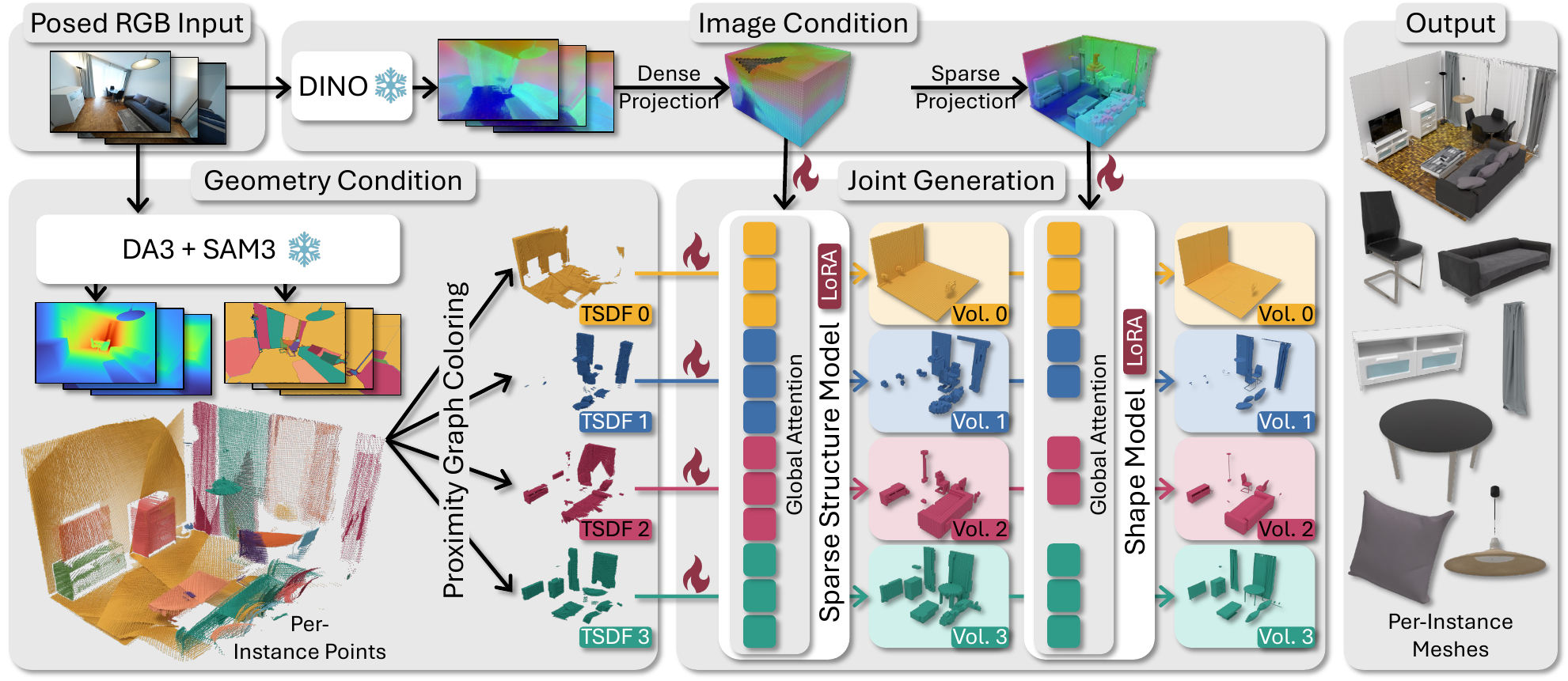}
  \caption{\textbf{Pipeline overview.} From the posed RGB input images, we predict depth and multi-view consistent instance masks, assign each instance to one of $N$ volumes, and fuse a per-volume TSDF as geometry condition (bottom left). Together with a 3D image condition lifted from projected image features (top), these guide a single model that generates all volumes jointly with global attention (bottom middle). Instances are extracted from each volume as separate PBR meshes (right).}
  \label{fig:pipeline}
\end{figure*}

\paragraph{Compositional reconstruction.}
Many recent methods compose scenes from individually obtained objects.
Each object is first generated with an object-level
prior~\cite{teamSAM3D3Dfy2025,tianSceneReGenGenerativeReconstruction2026,zadaianchukRecGenReconstructiveGeneration2026,siddiquiShapeRRobustConditional2026,niuWorldSculptGeneratingCompositional2026,yin3DFixerCoarsetoFineInplace2026,dongReplicateAnySceneZeroShotVideoto3D2026}, sometimes after completing its occluded view in image space~\cite{dogaruGeneralizable3DScene2024,sautter3DREGEN3DReconstruction2025,agarwalSceneCompleteOpenWorld3D2025}, or retrieved~\cite{huangLiteRealityGraphicsReady3D2025}, and then positioned.
Some methods predict pose with shape~\cite{teamSAM3D3Dfy2025,tianSceneReGenGenerativeReconstruction2026,zadaianchukRecGenReconstructiveGeneration2026}.
Others generate objects directly in a scene-anchored frame: ShapeR~\cite{siddiquiShapeRRobustConditional2026} conditions on the object's metric SLAM points, WorldSculpt~\cite{niuWorldSculptGeneratingCompositional2026} generates in a frame given by its 3D box and an anchor view, and 3D-Fixer~\cite{yin3DFixerCoarsetoFineInplace2026} completes objects in place on estimated geometry.
The rest place objects afterward via pose optimization against estimated geometry~\cite{dogaruGeneralizable3DScene2024,sautter3DREGEN3DReconstruction2025}, detected 3D boxes~\cite{huangLiteRealityGraphicsReady3D2025}, correspondences as in ReplicateAnyScene~\cite{dongReplicateAnySceneZeroShotVideoto3D2026}, scan alignment~\cite{agarwalSceneCompleteOpenWorld3D2025}, scene graphs~\cite{xuReSceneStructuredIndoor2026}, or editor feedback~\cite{qinLucidaParseGenerate2026}.
Poses and sometimes shapes may then be refined for physical plausibility~\cite{yaoCASTComponentAligned3D2025,maREST3DReconstructingPhysically2026,li$phi$ScenePhysicallyGrounded2026,leeSimuSceneSimulationReadyCompositional2026,xiaSimReconSimReadyCompositional2026}.
Throughout, each object's geometry is obtained independently, with inter-object relations entering only through
placement or refinement; placement errors can cause misaligned or intersecting objects, and the cost grows with
the number of objects.
We instead generate all objects jointly with their arrangement.

\paragraph{Joint reconstruction.}
Other methods model the scene as a whole.
Holistic generative methods produce complete scenes from images~\cite{schmid2026genreconbridginggenerativepriors}
or partial scans~\cite{mengSeen2SceneCompletingRealistic2026}, but without object identity.
Decomposed neural fields represent all objects jointly, adding overlap
constraints~\cite{wuObjectSDFImprovedObjectCompositional2023}, physical
plausibility~\cite{niPhyReconPhysicallyPlausible2024}, or per-object generative
completion~\cite{niDecompositionalNeuralScene2025}; others decompose scenes for segmentation and
editing~\cite{lyuTotaldecomDecomposed3d2024,yeGaussianGroupingSegment2024,zhuObjectGSObjectawareScene2025} or
complete objects with Gaussians~\cite{yangInstasceneComplete3d2025}.
Generative methods denoise one component per object or part with cross-component
attention~\cite{huangMIDIMultiInstanceDiffusion2024,mengSceneGenSingleImage3D2025,linPartCrafterStructured3D2025,liMoCAMixtureofComponentsAttention2025},
optionally plus an environment component~\cite{luoDistSceneObjecttoSceneDistillation2026}, condition instances on
scene context~\cite{lingIScene3DInstance2025}, or aggregate observations of related
objects~\cite{wuJRMJointReconstruction2026}.
These methods share context, but each object keeps its own representation, so the cost grows with the number of
objects; feed-forward regression avoids this but lacks a generative prior and was evaluated only on tabletop
scenes~\cite{duOCH3RObjectCentricHolistic2026}.
Closest to ours, PartPacker~\cite{tang2025partpacker} generates an object's parts in two volumes, placing touching
parts in different ones so that the cost is independent of the number of parts.
We extend this to $N$ volumes and entire scenes, with soft constraints for contacts that cannot all be separated.


%


\section{Method}
\label{sec:method}
Given an unordered set of $M$ RGB images $\mathcal{I}=\{I_k\}_{k=1}^{M}$ with known intrinsics $K_k$ and extrinsics $T_k$, our goal is to reconstruct the scene as a set of complete meshes
\begin{equation}
  \mathcal{S} \;=\; \{\mathcal{M}_s\} \,\cup\, \{\mathcal{M}_j\}_{j=1}^{J}
  \label{eq:output}
\end{equation}
in a common world frame, where $\mathcal{M}_s$ is the room structure (floor, walls, and ceiling) and
$\mathcal{M}_1,\dots,\mathcal{M}_J$ are the individual objects. The number of objects $J$ is not known in advance.

Our key idea is to distribute the objects over a small, fixed number $N$ of volumes such that touching objects are
kept apart (Sec.~\ref{sec:nvol}).
A single scene-level model built on TRELLIS.2 and GenRecon (Sec.~\ref{sec:prelim}) generates all volumes jointly
(Sec.~\ref{sec:joint}), conditioned on the images and a per-volume geometry condition (Sec.~\ref{sec:cond}), and
covers large scenes with chunks (Sec.~\ref{sec:inference}). 
Fig.~\ref{fig:pipeline} gives an overview.


\subsection{Preliminaries}
\label{sec:prelim}

TRELLIS.2~\cite{xiang2025trellis2} generates a 3D asset in three conditional flow-matching stages: a sparse structure stage generates a dense latent grid that decodes to a coarse voxel occupancy; a shape stage generates a structured latent on the occupied voxels, which decodes to a mesh; and a texture stage generates a second structured latent on the same voxels, conditioned on the shape latent, which decodes to PBR attributes.

GenRecon~\cite{schmid2026genreconbridginggenerativepriors} extends TRELLIS.2 from objects to multi-view scene reconstruction. 
It lifts DINO~\cite{siméoni2025dinov3} features from the posed views into a 3D feature grid aligned with the latent grid and covers large scenes using spatially localized chunks. 
Its output is a single mesh without object identity: where two objects touch, only one surface exists between them, so they cannot be separated into complete instances.


\subsection{$N$-Volume Scene Representation}
\label{sec:nvol}
\begin{figure*}[t]
  \centering
  \includegraphics[width=\textwidth]{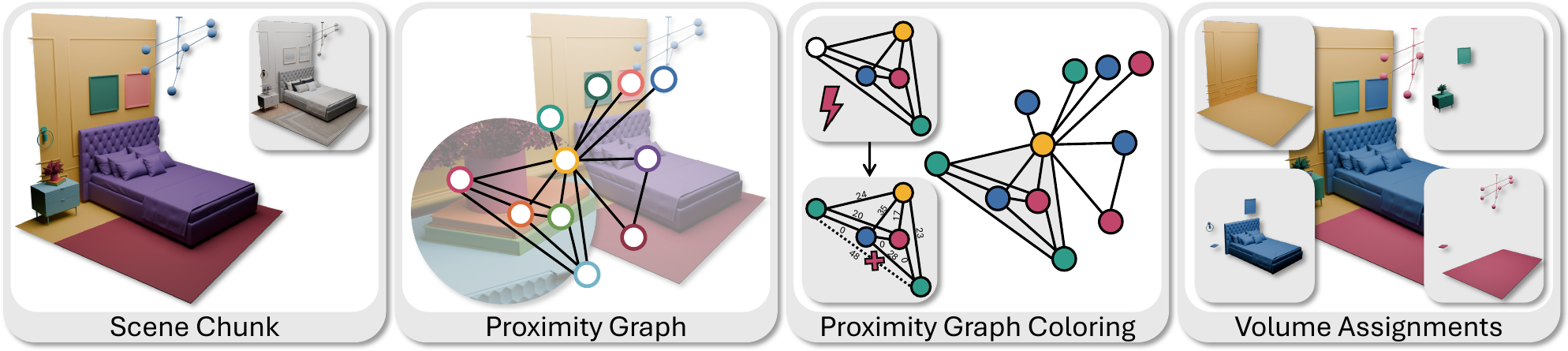}
  \caption{\textbf{Instance-to-volume assignment at training time.} Given a scene chunk with per-instance meshes (1), we build a proximity graph, where all instances close to each other are connected (2). Using the pairwise mesh distances $d_{ij}$, we solve a weighted minimum $N$-coloring problem that encourages only the most distant instance pairs to share the same color (3). Based on these colorings, we divide the instances into the $N$ co-located volumes (4).}
  \label{fig:volume_assign}
\end{figure*}
Generating each object individually yields complete objects, but costs grow with their number, and the objects are generated without scene context.
Since objects need to be kept apart only where they touch, it suffices to distribute them over a few volumes such that no two touching objects share one: a graph-coloring problem with objects as nodes, contacts as edges, and volumes as colors.
PartPacker~\cite{tang2025partpacker} uses this idea to pack the parts of a single object into two volumes; we extend it to $N$ volumes and entire scenes, and relax it where no valid coloring exists.

\paragraph{Representation.}
We represent a scene by $N$ co-located volumes $\mathcal{X}_0,\dots,\mathcal{X}_{N-1}$: all volumes cover the same region of 3D space with identical position, orientation, and extent, and differ only in which objects they contain.
An assignment $c_j\in\{0,\dots,N{-}1\}$ places every element $j\in\{s,1,\dots,J\}$ of $\mathcal{S}$ in exactly one
volume, so that each volume is the union of its meshes, $\mathcal{X}_v=\bigcup_{j:\,c_j=v}\mathcal{M}_j$.
The room structure always lies in volume~$0$, and it can also contain objects that do not touch the structure. 
The number of volumes is fixed (we use $N{=}4$), but the number of objects per volume is not, so the generation cost does not grow with the number of objects in the scene.

\paragraph{Volume assignment.}
We build a proximity graph $G=(\mathcal{V},\mathcal{E})$ over the objects and a single structure node $s$, into which all structural elements are merged. 
Two nodes are connected if they are close in 3D space, with a positive weight $w_{ij}$ that measures their proximity.
We then solve the weighted minimum-conflict $N$-coloring
\begin{equation}
  c^\ast \;=\; \argmin_{c\,:\;c_s=0}\;\; \sum_{(i,j)\in\mathcal{E}} w_{ij}\,\mathds{1}\!\left[c_i=c_j\right]
  \label{eq:coloring}
\end{equation}
with CP-SAT~\cite{cpsat}.
The constraints are soft because a conflict-free coloring does not always exist: 3D contact graphs need not be planar.
In such cases, Eq.~\eqref{eq:coloring} places the conflicts on the pairs with the smallest total weight, so touching pairs stay separated whenever possible (Fig.~\ref{fig:volume_assign}). 
Only the graph construction differs between training and inference:
For training, nodes are connected if their minimum surface distance $d_{ij}$ is below a threshold $\tau$, with $w_{ij}=1+\lambda\,(1-d_{ij}/\tau)$ and a large $\lambda$, so that touching pairs are the most expensive to merge. 
For inference, each instance is lifted with the depth into a voxel set $V_i$, and $w_{ij}=\big|(V_i\oplus B_r)\cap(V_j\oplus B_r)\big|$ counts the overlap of the sets dilated by a ball $B_r$, which is robust to noise in the lifted points.

\paragraph{Instance extraction.}
Since the objects within a volume are separated, we recover $\hat{\mathcal{S}}$ from the generated volume meshes $\hat{\mathcal{X}}_v$ by connected components. 
We voxelize each mesh surface at a fine resolution, label it with 26-connectivity, merge components below a minimum surface area into a nearby large component or discard them as floaters, and assign every face to its nearest kept voxel. 
The largest component of volume~$0$ is the room shell $\hat{\mathcal{M}}_s$. 

\subsection{Joint N-Volume Generation}
\label{sec:joint}

We fine-tune all three TRELLIS.2~\cite{xiang2025trellis2} stages to jointly generate the $N$ volumes. In every stage, we concatenate the tokens of all volumes into a single sequence: in the sparse structure stage, the flattened dense latent grids $\mathbf{z}_v$ (length $N\cdot R^3$); in the shape and texture stages, the sparse tokens on the occupied voxels of every volume. 
Since tokens at the same position $\mathbf{p}$ in different volumes share their positional encoding, we add a learnable volume embedding $e_v$ to each token.
Global attention across all tokens keeps the volumes spatially consistent and lets them agree on which volume
generates which object.
Each stage is fine-tuned with LoRA~\cite{hu2021loralowrankadaptationlarge} on the attention and MLP layers; only the adapters, the volume embeddings, and the conditioning modules are trained.

\subsection{Volume-Aware Conditioning}
\label{sec:cond}
The model receives an image condition $\mathbf{f}$, shared by all volumes, and in the sparse structure stage, a geometry condition $\mathbf{g}_v$ per volume. 
Both are feature grids aligned with the latent grid.
Before every transformer block, each passes through its own zero-initialized linear layer and is added to the token at the same grid position.

\paragraph{3D image condition.}
Following GenRecon~\cite{schmid2026genreconbridginggenerativepriors}, we project DINO~v3~\cite{siméoni2025dinov3} features of all views into 3D, but onto a grid twice as fine as the latent grid. 
The features of each fine voxel are aggregated across views using learned weights in the style of IBRNet~\cite{wang2021ibrnet}, and a small 3D CNN then reduces the grid to the latent resolution. 
All three stages use this condition, and the image cross-attention of TRELLIS.2 receives an empty context.

\paragraph{3D geometry condition.}
Since $\mathbf{f}$ is identical for all volumes, many assignments explain it equally well, and without further cues the model struggles to commit to one: objects appear in two volumes, are split across volumes, or are missing
(Sec.~\ref{sec:ablation}).  
The sparse structure stage therefore also receives the intended assignment: we lift the instance masks with the
per-view metric depth into per-instance point clouds and assign them to volumes with Eq.~\eqref{eq:coloring}.
Following this assignment, we fuse one TSDF $\mathcal{T}_v$ per volume from the depth pixels of its instances, while the pixels of all other volumes only contribute free space, since an object of volume $v$ may be hidden behind them. 
Pixels without a volume (uncertain mask borders, and undetected objects) are fused into an additional unassigned TSDF $\mathcal{T}_\varnothing$, so that their geometry is not lost.
A small 3D CNN, shared across volumes, encodes $[\mathcal{T}_v,\mathcal{T}_\varnothing]$ into $\mathbf{g}_v$.
During training, the assignment is derived from the ground-truth meshes (Sec.~\ref{sec:nvol}) and applied to ground-truth instance masks, whereas the depth is predicted by the same estimator as at inference time, so the 3D geometry condition includes realistic depth errors. 
To make the model robust to imperfect predicted masks, we corrupt the ground-truth masks with the error types we observe in predictions, such as missed or split instances, boundary jitter, spurious masks, and wrong volumes.

\paragraph{Two-scale guidance.}
We drop $\mathbf{f}$ and $\mathbf{g}$ independently during training, which enables classifier-free guidance~\cite{ho2022classifierfreediffusionguidance} with one scale per condition, as in InstructPix2Pix~\cite{brooks2023instructpix2pixlearningfollowimage}:
\begin{equation}
  \hat{v} \;=\; v_\varnothing \;+\; s_{\text{img}}\,\big(v_{\mathbf{f}} - v_\varnothing\big)
  \;+\; s_{\text{geo}}\,\big(v_{\mathbf{f},\mathbf{g}} - v_{\mathbf{f}}\big),
  \label{eq:guidance}
\end{equation}
where $v_\varnothing$, $v_{\mathbf{f}}$, and $v_{\mathbf{f},\mathbf{g}}$ are the predictions without condition, with the image condition, and with both conditions.

\subsection{Scene-Scale Inference}
\label{sec:inference}
\paragraph{Predicted inputs.}
At inference time, we predict depth for all views jointly with Depth Anything 3~\cite{depthanything3} given the camera poses, detect instances per view with SAM3~\cite{carion2026sam3segmentconcepts}, and match masks across views by reprojecting their lifted points. 
We compute the assignment (Sec.~\ref{sec:cond}) once for the whole scene, so all chunks share it. 
Objects without a mask remain unassigned, and the model determines their volume on its own.

\paragraph{Chunked sampling with random shifts.}
As in GenRecon, the model operates on chunks, cubes spanning the room height that tile the scene in $x$ and $y$, and all chunks are denoised together on one scene-wide latent. 
GenRecon averages overlapping chunks in the style of MultiDiffusion~\cite{bar2023multidiffusion}, which causes holes and cracks, since averaging TRELLIS.2 latents does not average the decoded geometry, and splits objects
straddling a seam between volumes.
Inspired by SpotDiffusion~\cite{frolov2025spotdiffusionfastapproachseamless}, we instead tile the scene at every denoising step with a single grid of non-overlapping chunks, shifted by a random offset in $x$ and $y$. 
Every voxel takes its prediction from exactly one chunk, and since the seams move between steps, none persists over the trajectory.
All three stages are sampled this way.

\section{Experiments}
\label{sec:experiments}

\begin{figure*}[ht]
  \centering
  \newcommand{\qualrow}[1]{%
    \includegraphics[width=\linewidth]{figures/qualitative/#1_mv3dis.jpg} &
    \includegraphics[width=\linewidth]{figures/qualitative/#1_shaper.jpg} &
    \includegraphics[width=\linewidth]{figures/qualitative/#1_astra.jpg} &
    \includegraphics[width=\linewidth]{figures/qualitative/#1_ours_70k.jpg} &
    \includegraphics[width=\linewidth]{figures/qualitative/#1_gt.jpg} \\
  }
  \setlength{\tabcolsep}{1pt}
  \renewcommand{\arraystretch}{0.5}
  \begin{tabular}{*{5}{>{\centering\arraybackslash}p{0.192\textwidth}}}
    \small GenRecon+MV3DIS & \small ShapeR & \small Astra & \small Ours & \small Ground truth \\[3pt]
    \qualrow{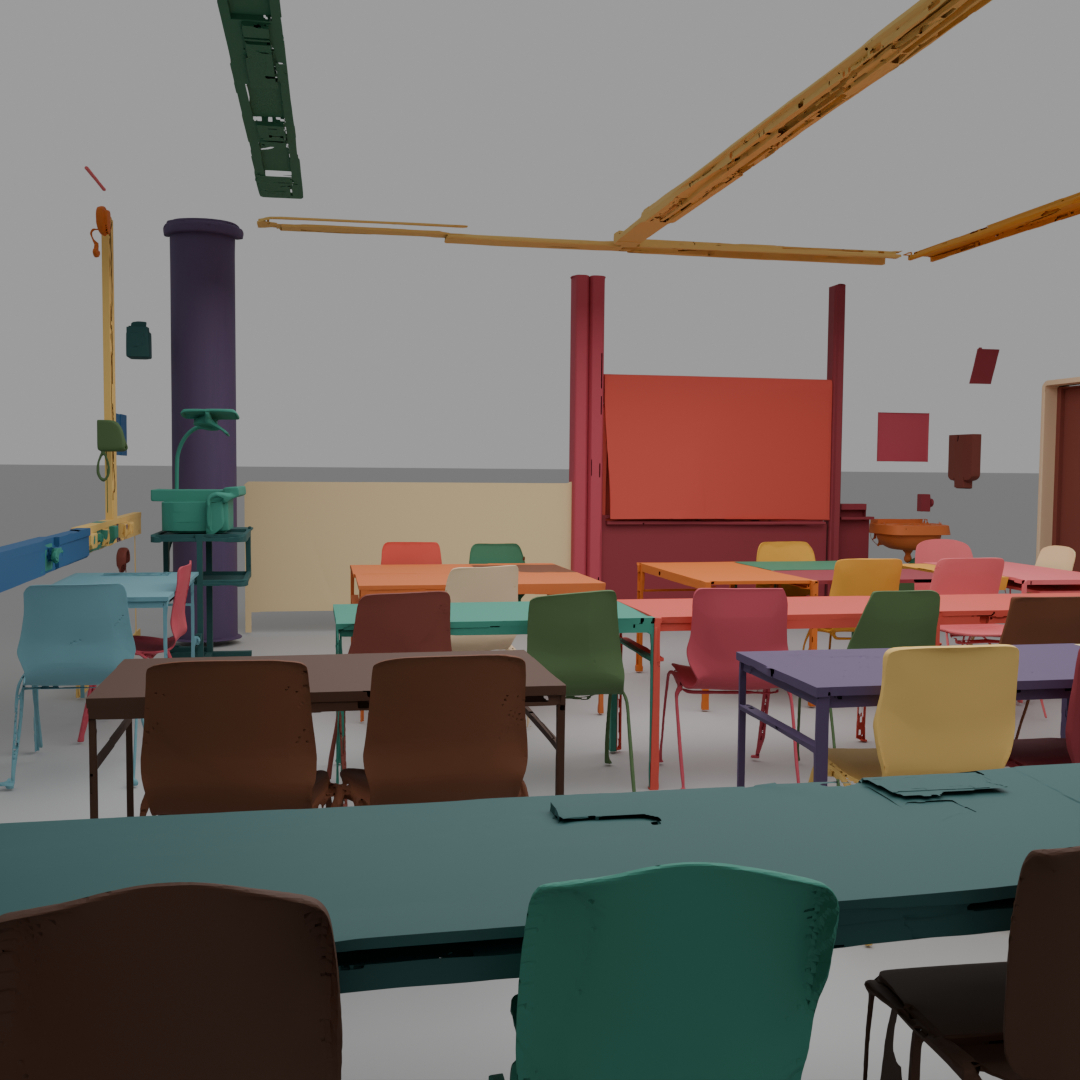}
    \qualrow{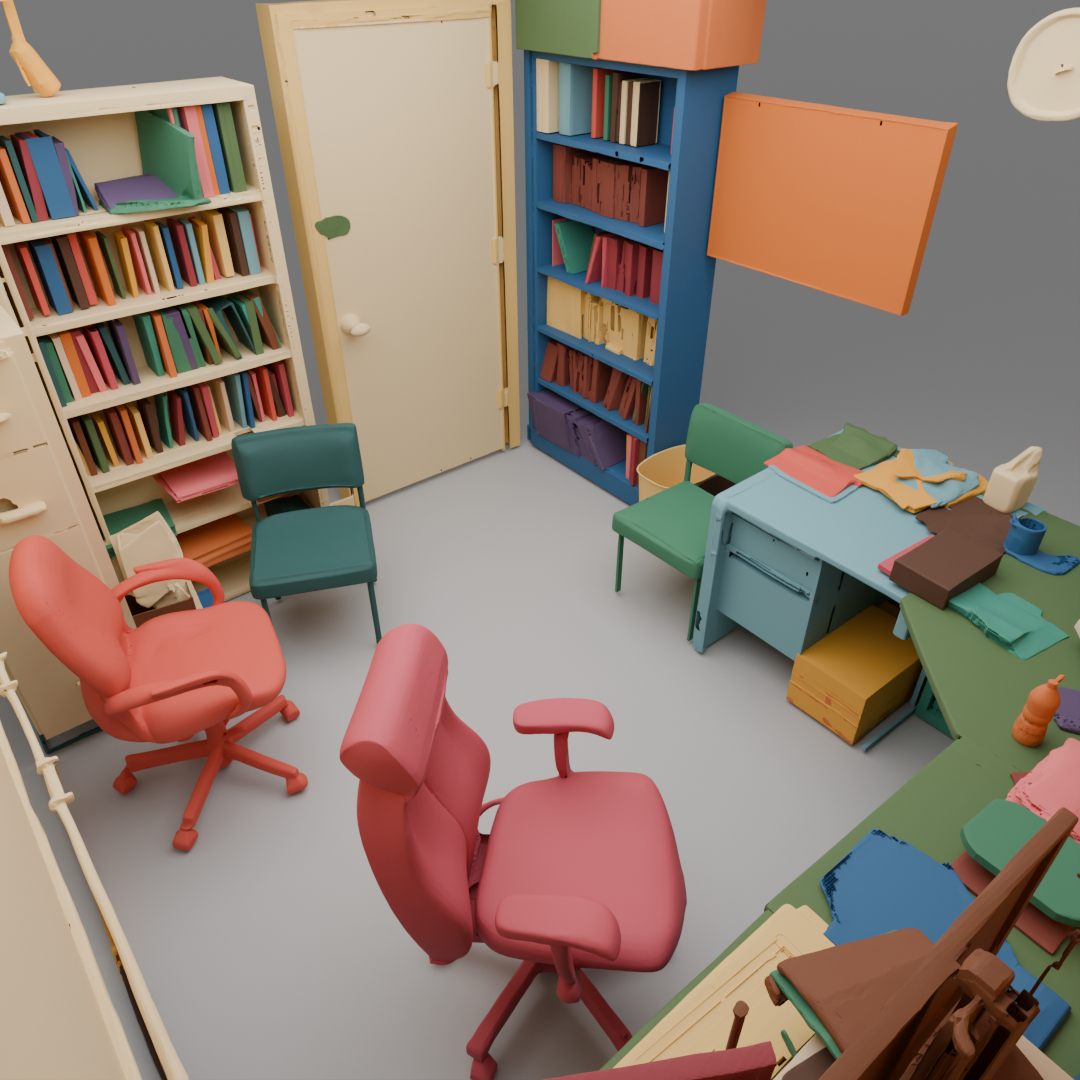}
    \qualrow{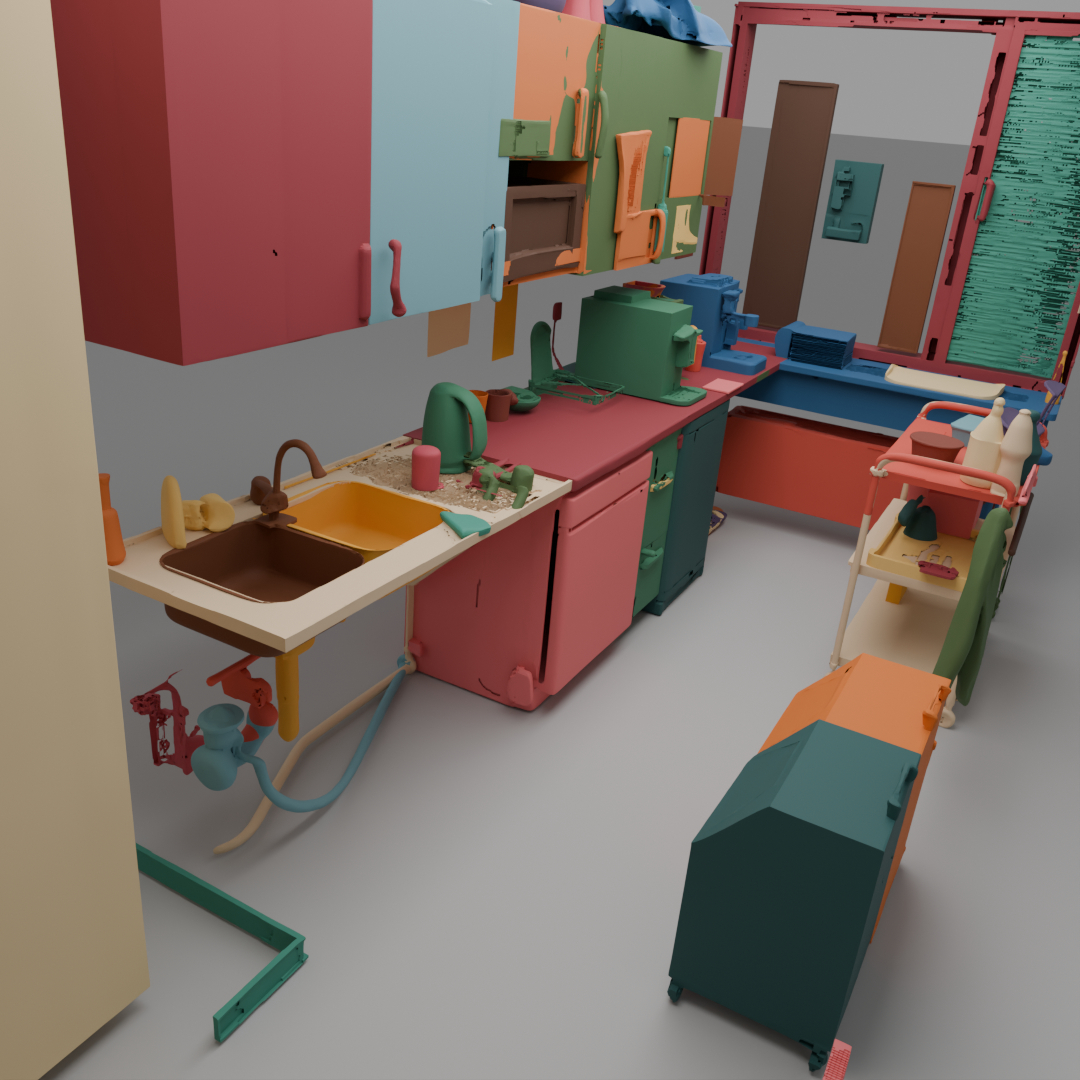}
    \qualrow{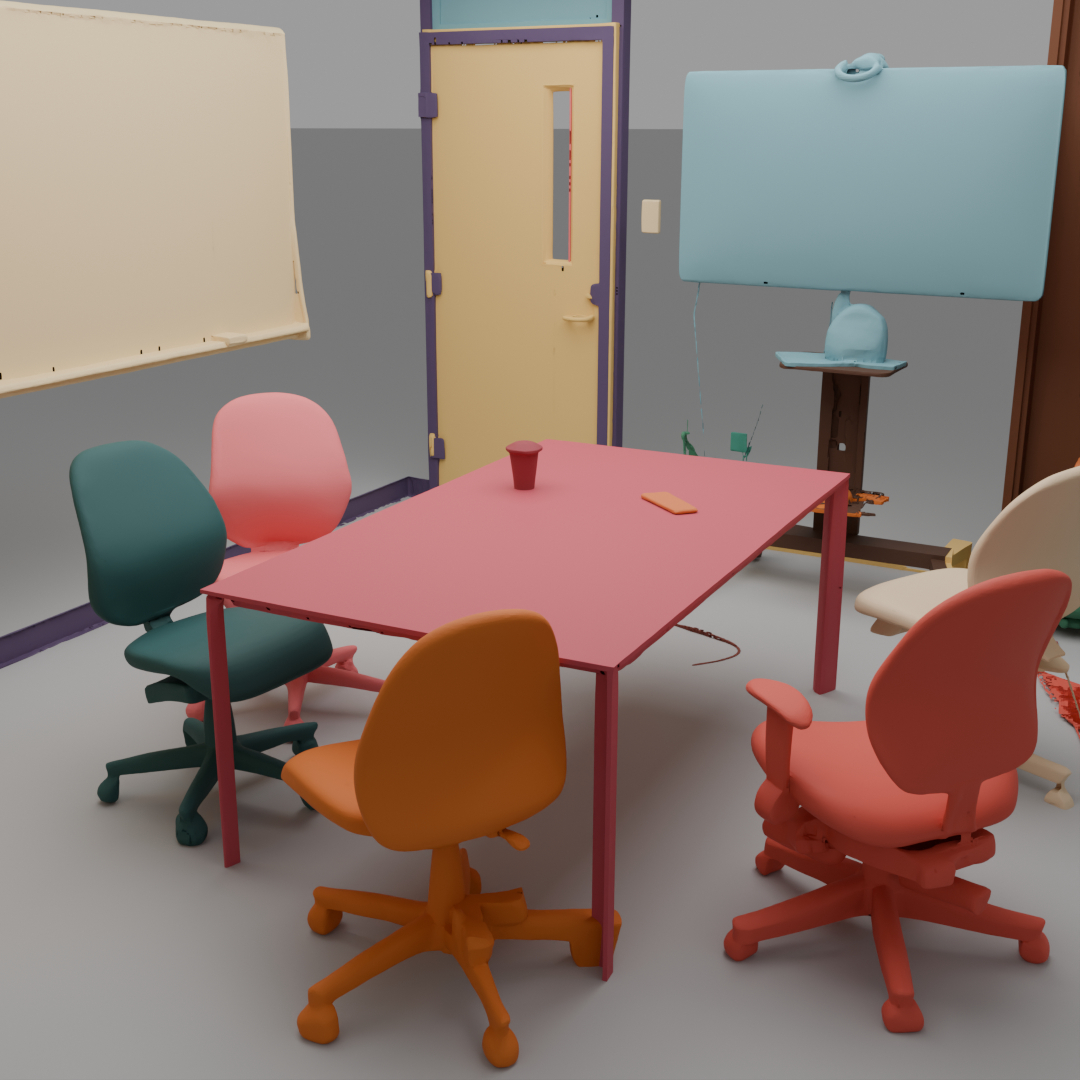}
  \end{tabular}
  \caption{\textbf{Qualitative evaluation on real-world data.} Reconstruction and decomposition performance from eight input images and predicted depth and instance masks on scenes from ScanNet++, with structural elements removed. Compared with the baselines, our results more closely match the ground truth scan.}
  \label{fig:qual_comp}
\end{figure*}

\begin{table*}[t]
  \centering
  \caption{\textbf{Quantitative evaluation on real-world data.} We evaluate 3D reconstruction and decomposition metrics on 20 scenes from ScanNet++ given eight input images and predicted depth and instance masks.}
  \label{tab:scannetpp_objects}
  \small
  \setlength{\tabcolsep}{3pt}
  \begin{tabular}{lrrrrrrr@{\hspace{10pt}}rrrrr}
    \toprule
     & \multicolumn{7}{c}{Reconstruction} & \multicolumn{5}{c}{Decomposition} \\
    \cmidrule(lr){2-8} \cmidrule(lr){9-13}
     & \multirow{2}{*}{CD (cm) $\downarrow$} & \multicolumn{5}{c}{F-score $\uparrow$} & \multirow{2}{*}{NC $\uparrow$} & \multicolumn{5}{c}{F1 $\uparrow$} \\
    \cmidrule(lr){3-7} \cmidrule(lr){9-13}
    Method & & @1 & @2 & @4 & @8 & @16 & & @1 & @2 & @4 & @8 & @16 \\
    \midrule
    GenRecon+SpaCeFormer & 16.07 & 0.074 & 0.190 & 0.368 & 0.550 & 0.708 & 0.771 & 0.001 & 0.017 & 0.108 & 0.288 & 0.435 \\
    GenRecon+MV3DIS & \cellcolor{yellow!40}11.46 & 0.097 & 0.234 & 0.444 & 0.666 & 0.820 & \cellcolor{yellow!40}0.790 & 0.001 & 0.018 & 0.139 & 0.331 & 0.465 \\
    \midrule
    ReplicateAnyScene & 14.02 & 0.067 & 0.171 & 0.338 & 0.559 & 0.773 & 0.743 & 0.003 & 0.045 & 0.203 & 0.403 & 0.570 \\
    WorldSculpt & 19.32 & 0.090 & 0.209 & 0.358 & 0.506 & 0.640 & 0.731 & \cellcolor{yellow!40}0.009 & \cellcolor{yellow!40}0.084 & \cellcolor{yellow!40}0.233 & 0.395 & 0.519 \\
    ShapeR & 11.79 & \cellcolor{orange!35}0.124 & \cellcolor{orange!35}0.295 & \cellcolor{yellow!40}0.520 & \cellcolor{yellow!40}0.719 & \cellcolor{yellow!40}0.832 & \cellcolor{orange!35}0.796 & \cellcolor{orange!35}0.017 & \cellcolor{orange!35}0.095 & \cellcolor{orange!35}0.285 & \cellcolor{orange!35}0.484 & \cellcolor{yellow!40}0.590 \\
    \midrule
    Astra & \cellcolor{orange!35}7.37 & \cellcolor{yellow!40}0.123 & \cellcolor{yellow!40}0.273 & \cellcolor{orange!35}0.523 & \cellcolor{orange!35}0.769 & \cellcolor{orange!35}0.909 & 0.777 & 0.005 & 0.052 & 0.216 & \cellcolor{yellow!40}0.479 & \cellcolor{red!30}\textbf{0.634} \\
    \midrule
    Ours & \cellcolor{red!30}\textbf{5.61} & \cellcolor{red!30}\textbf{0.226} & \cellcolor{red!30}\textbf{0.504} & \cellcolor{red!30}\textbf{0.743} & \cellcolor{red!30}\textbf{0.869} & \cellcolor{red!30}\textbf{0.926} & \cellcolor{red!30}\textbf{0.818} & \cellcolor{red!30}\textbf{0.027} & \cellcolor{red!30}\textbf{0.164} & \cellcolor{red!30}\textbf{0.412} & \cellcolor{red!30}\textbf{0.554} & \cellcolor{orange!35}0.624 \\
    \bottomrule
  \end{tabular}
\end{table*}
\begin{table*}[t]
  \centering
  \caption{\textbf{Quantitative evaluation on synthetic data.} We evaluate 3D reconstruction and decomposition metrics on 20 scenes from SceneSmith given eight input images and predicted depth and instance masks.}
  \label{tab:scenesmith_objects}
  \small
  \setlength{\tabcolsep}{3pt}
  \begin{tabular}{lrrrrrrr@{\hspace{10pt}}rrrrr}
    \toprule
     & \multicolumn{7}{c}{Reconstruction} & \multicolumn{5}{c}{Decomposition} \\
    \cmidrule(lr){2-8} \cmidrule(lr){9-13}
     & \multirow{2}{*}{CD (cm) $\downarrow$} & \multicolumn{5}{c}{F-score $\uparrow$} & \multirow{2}{*}{NC $\uparrow$} & \multicolumn{5}{c}{F1 $\uparrow$} \\
    \cmidrule(lr){3-7} \cmidrule(lr){9-13}
    Method & & @1 & @2 & @4 & @8 & @16 & & @1 & @2 & @4 & @8 & @16 \\
    \midrule
    GenRecon+SpaCeFormer & 22.96 & 0.113 & 0.230 & 0.367 & 0.501 & 0.635 & 0.643 & 0.003 & 0.058 & 0.203 & 0.344 & 0.469 \\
    GenRecon+MV3DIS & 12.04 & 0.138 & 0.283 & 0.444 & 0.615 & 0.790 & 0.697 & 0.009 & 0.085 & 0.233 & 0.387 & 0.526 \\
    \midrule
    ReplicateAnyScene & 10.19 & 0.203 & 0.370 & 0.571 & 0.763 & 0.868 & 0.720 & 0.081 & 0.211 & 0.440 & 0.599 & \cellcolor{yellow!40}0.671 \\
    WorldSculpt & 11.30 & \cellcolor{orange!35}0.264 & 0.438 & 0.600 & 0.738 & 0.836 & \cellcolor{yellow!40}0.747 & \cellcolor{orange!35}0.134 & \cellcolor{orange!35}0.303 & 0.435 & 0.528 & 0.591 \\
    ShapeR & \cellcolor{yellow!40}9.90 & \cellcolor{yellow!40}0.257 & \cellcolor{orange!35}0.463 & \cellcolor{yellow!40}0.639 & \cellcolor{yellow!40}0.784 & \cellcolor{yellow!40}0.880 & 0.744 & \cellcolor{yellow!40}0.088 & 0.271 & \cellcolor{yellow!40}0.473 & \cellcolor{yellow!40}0.608 & 0.664 \\
    \midrule
    Astra & \cellcolor{orange!35}8.63 & 0.254 & \cellcolor{yellow!40}0.450 & \cellcolor{orange!35}0.673 & \cellcolor{orange!35}0.847 & \cellcolor{orange!35}0.919 & \cellcolor{orange!35}0.751 & 0.075 & \cellcolor{yellow!40}0.301 & \cellcolor{orange!35}0.611 & \cellcolor{red!30}\textbf{0.772} & \cellcolor{red!30}\textbf{0.839} \\
    \midrule
    Ours & \cellcolor{red!30}\textbf{5.07} & \cellcolor{red!30}\textbf{0.456} & \cellcolor{red!30}\textbf{0.692} & \cellcolor{red!30}\textbf{0.835} & \cellcolor{red!30}\textbf{0.908} & \cellcolor{red!30}\textbf{0.945} & \cellcolor{red!30}\textbf{0.791} & \cellcolor{red!30}\textbf{0.230} & \cellcolor{red!30}\textbf{0.502} & \cellcolor{red!30}\textbf{0.666} & \cellcolor{orange!35}0.746 & \cellcolor{orange!35}0.782 \\
    \bottomrule
  \end{tabular}
\end{table*}

\subsection{Experiment Setup}
\label{sec:setup}

\paragraph{Implementation details.}
We train on 10k synthetic indoor scenes: 5k from SAGE-10k~\cite{xia2026sagescalableagentic3d}, 2k single-room
scenes generated with Infinigen Indoors~\cite{infinigen2024indoors}, and 3k from Physicl~\cite{liu2026physiclscenesartistcreated3d}.
From each scene, we sample five chunks for training, render views with ground-truth instance masks, and predict their depth with Depth Anything~3~\cite{depthanything3}.
We fine-tune the TRELLIS.2~\cite{xiang2025trellis2} sparse structure model at resolution $64^3$ and the shape and
texture models at resolution $512^3$, using LoRA~\cite{hu2021loralowrankadaptationlarge} with rank $32$ and $\alpha{=}64$.
We optimize with AdamW~\cite{loshchilov2019decoupledweightdecayregularization} at a batch size of 32, with learning
rates of $3\cdot10^{-5}$ for the LoRA adapters and $10^{-4}$ for the new modules.
The sparse structure model is trained on eight GPUs with 80\,GB of memory each, and the shape and texture models on four.

\paragraph{Evaluation.}
We evaluate on 20 single-room scenes from SceneSmith~\cite{scenesmith2026}, randomly selected across its difficulty categories, and 20 real captures from the ScanNet++ validation split~\cite{yeshwanthliu2023scannetpp}.
SceneSmith is unseen during training and provides complete object geometry, including occluded and contact
surfaces, whereas the scanned ground truth of ScanNet++ covers observed surfaces only.
Each scene uses eight input images with ground-truth poses; depth and instance masks are predicted as in
Sec.~\ref{sec:inference}, with SAM3~\cite{carion2026sam3segmentconcepts} prompted by the ground-truth category names.
We exclude walls, floor, and ceiling from all evaluations, since not all baselines reconstruct them.

\paragraph{Metrics.}
We evaluate reconstruction and decomposition quality in 3D.
On ScanNet++, we evaluate only observed surfaces by casting the rays of the ground-truth scans and comparing each scanned point with the first hit on the prediction; on SceneSmith, we densely sample both surfaces.
For reconstruction, we compare the union of all predicted object meshes with the union of all ground-truth objects and report Chamfer distance (CD), F-score, and normal consistency (NC).
For decomposition, we match predicted to ground-truth objects with a one-to-one Hungarian assignment that maximizes
$\sum (\theta - d_{ab})$ over pairs with symmetric Chamfer distance $d_{ab} < \theta$.
Matched pairs are true positives, unmatched predictions false positives, and unmatched ground-truth objects false negatives; we report the F1-score for $\theta \in \{1, 2, 4, 8, 16\}$\,cm.

\paragraph{Baselines.}
We compare against three groups of methods.
\emph{Reconstruct-then-segment:} we reconstruct the scene with the sparse-view method GenRecon~\cite{schmid2026genreconbridginggenerativepriors} and segment the resulting mesh into instances, either directly in 3D with SpaceFormer~\cite{choy2026spaceformer} or by lifting multi-view 2D masks with MV3DIS~\cite{zhao2026mv3dismultiviewmaskmatching}.
\emph{Compositional reconstruction:} ShapeR~\cite{siddiquiShapeRRobustConditional2026}, WorldSculpt~\cite{niuWorldSculptGeneratingCompositional2026}, and ReplicateAnyScene~\cite{dongReplicateAnySceneZeroShotVideoto3D2026} generate each object separately and compose them into a scene.
\emph{Agentic modeling:} We compare against Astra~\cite{openai2026gpt6astra}, an LLM agent that models the scene in Blender.
For a fair comparison, all methods receive the same input views, and methods that take instance masks or depth use the same predicted masks and depth as ours.


\subsection{Experimental Results}

\paragraph{Real-world decomposed scene reconstruction.}
On ScanNet++ (Tab.~\ref{tab:scannetpp_objects}), our method outperforms all baselines on every reconstruction metric and on all but the loosest decomposition threshold, where it is on par with Astra.
The margin is largest at tight thresholds and shrinks at loose ones, indicating that the baselines recover roughly the right objects but place them imprecisely, since each object is generated independently and positioned afterward.
By generating all objects jointly in a common frame, our method keeps them aligned with each other and with the
observations, as Fig.~\ref{fig:qual_comp} shows: baseline objects are often misaligned or intersect, whereas ours
closely match the ground truth.
Although trained only on synthetic data, our method transfers well to real captures.

\begin{figure*}[t]
  \centering
  \setlength{\tabcolsep}{1pt}
  \newcommand{\litw}{\dimexpr(\textwidth-8\tabcolsep)/5\relax}
  \newcommand{\litrow}[1]{%
    \includegraphics[width=\litw]{figures/lit/#1_input.jpg} &
    \includegraphics[width=\litw]{figures/lit/#1_decomp.jpg} &
    \includegraphics[width=\litw]{figures/lit/#1_normal.jpg} &
    \includegraphics[width=\litw]{figures/lit/#1_pbr_light1.jpg} &
    \includegraphics[width=\litw]{figures/lit/#1_pbr_light2.jpg} \\
  }
  \renewcommand{\arraystretch}{0.5}
  \begin{tabular}{@{}ccccc@{}}
    \small Input image & \small Decomposition & \small Normals & \small Lit v1 & \small Lit v2 \\[3pt]
    \litrow{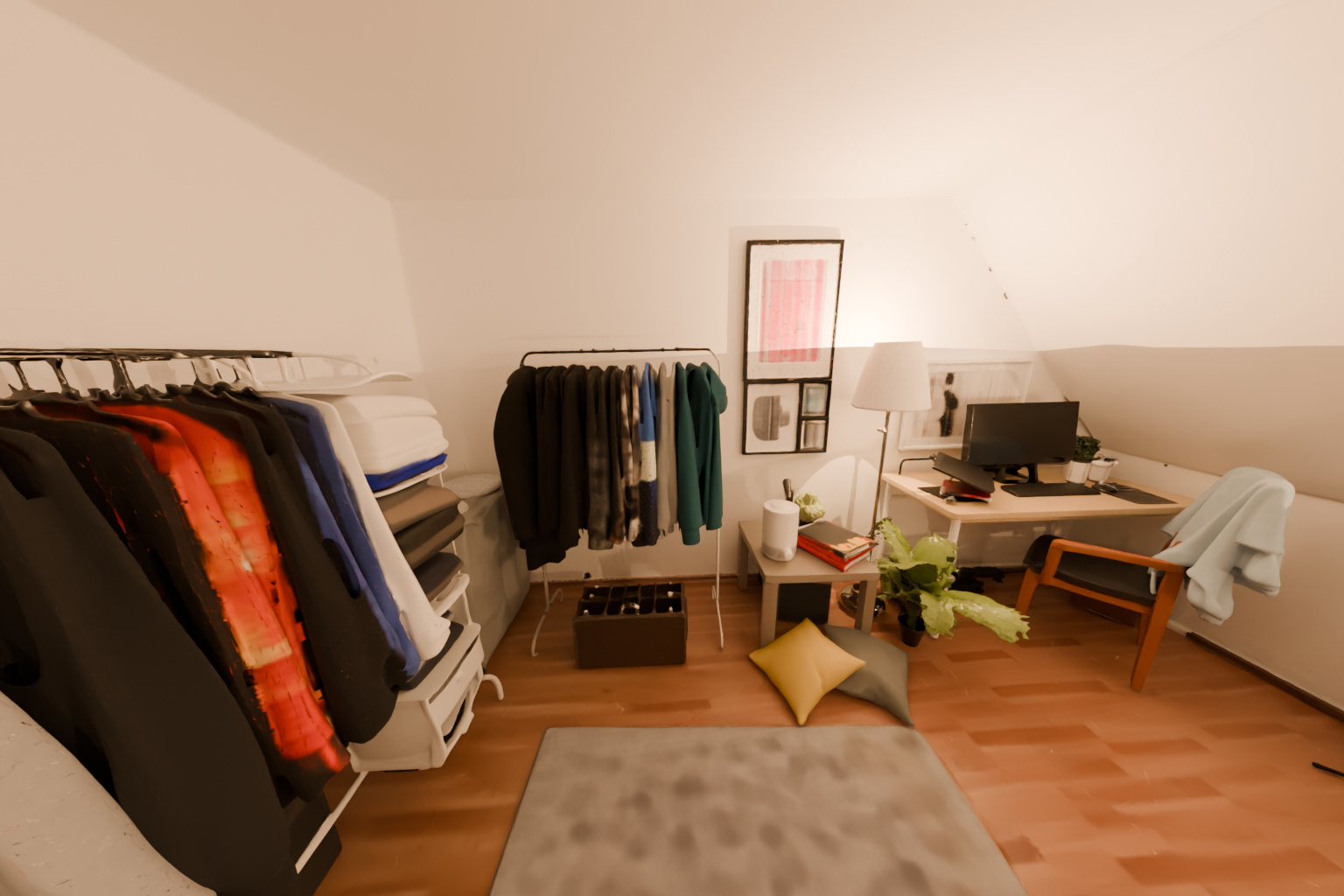}
    \litrow{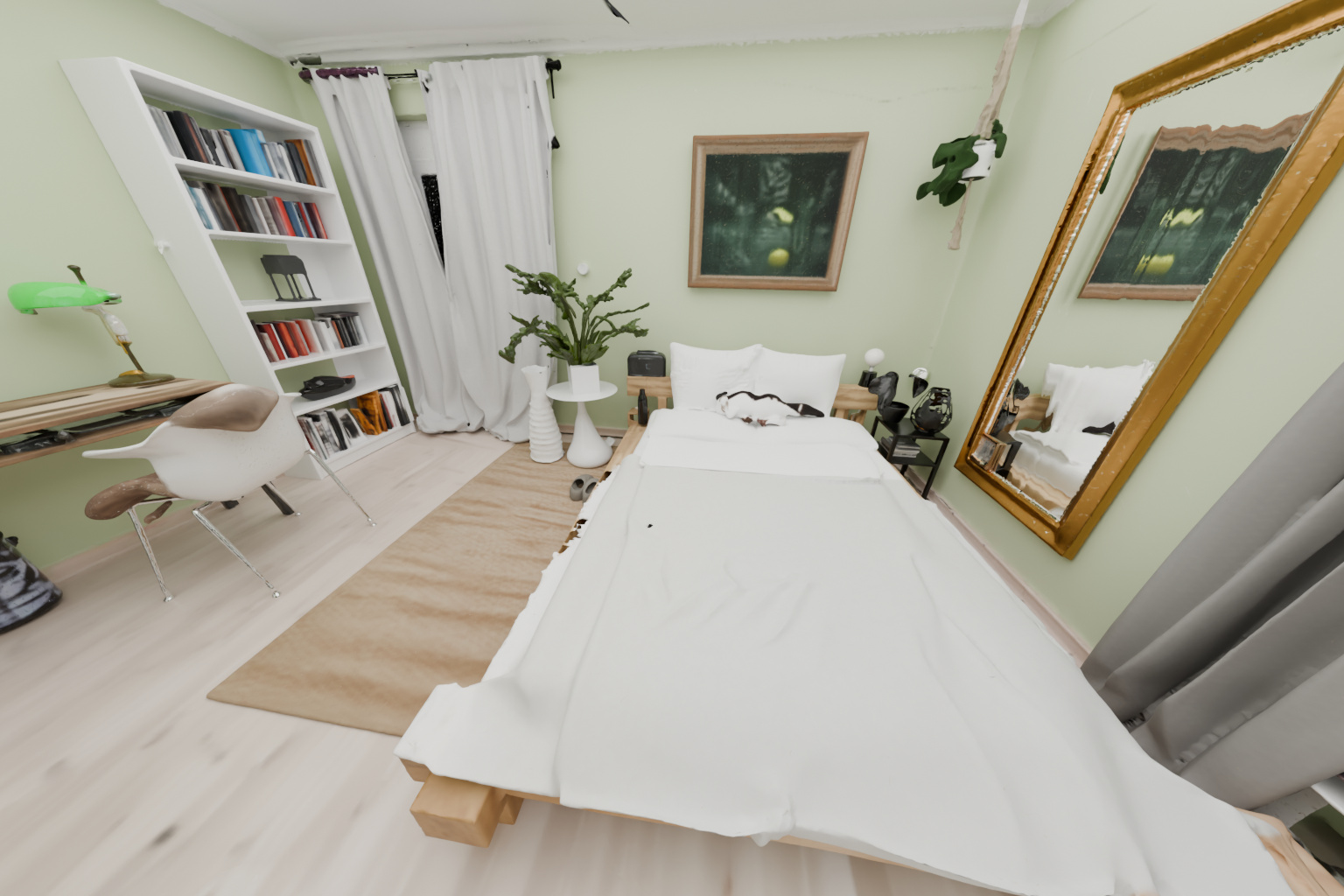}
    \litrow{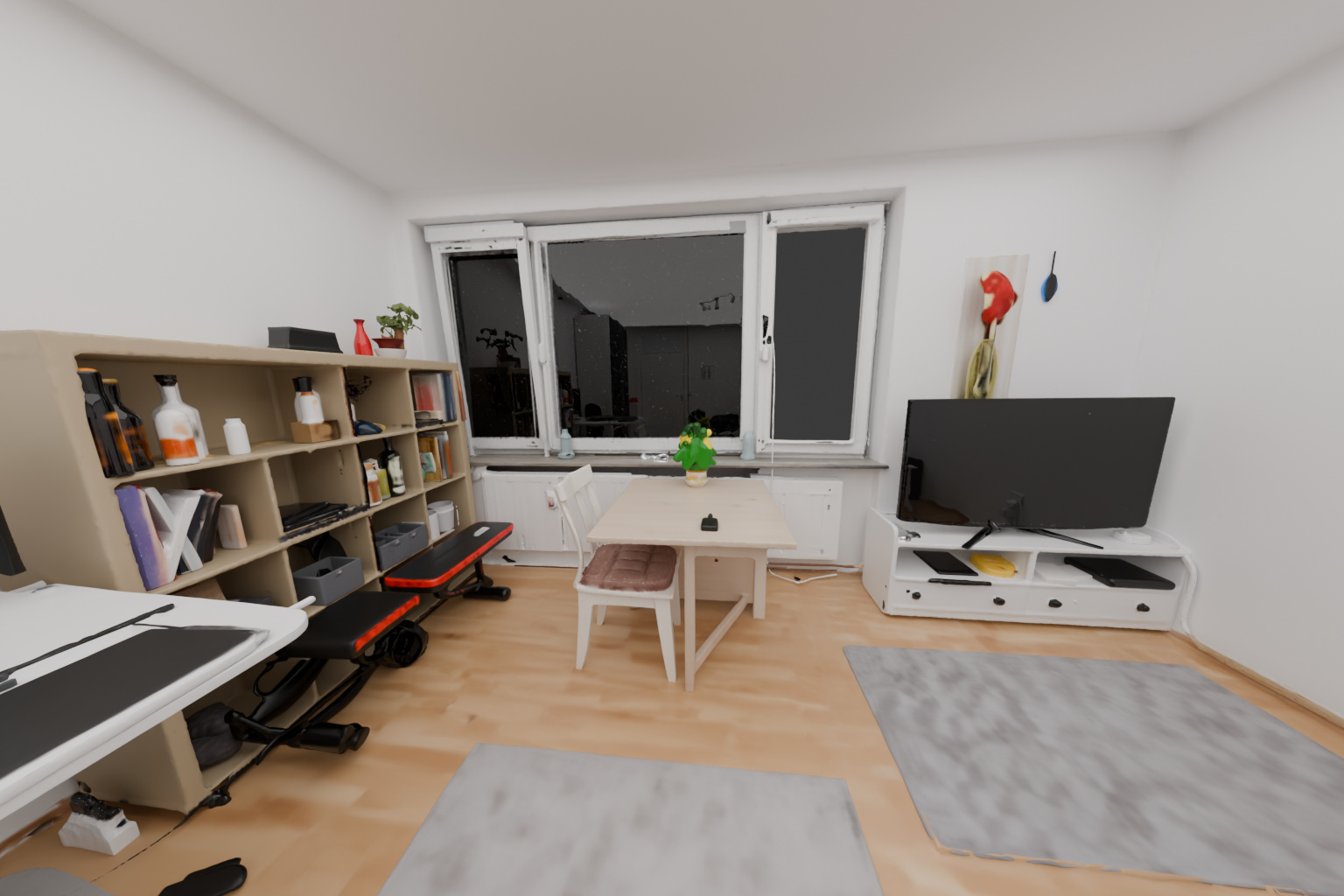}
    \litrow{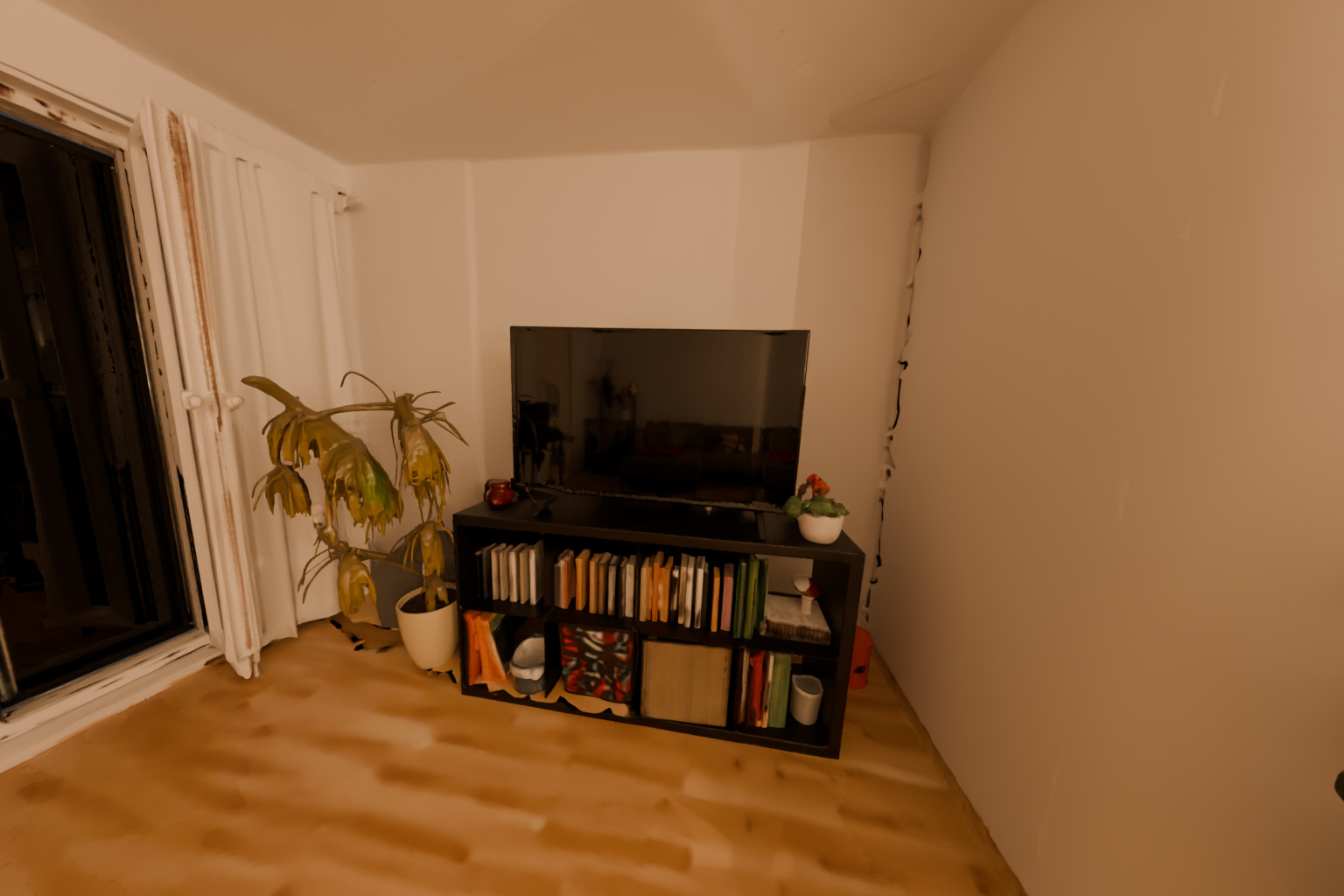}
  \end{tabular}
  \caption{\textbf{Qualitative decomposition, reconstruction, and relighting results on real-world data.} Results on ScanNet++ from eight input images and predicted depth and instance masks. Our approach reconstructs highly realistic relightable textures.}
  \label{fig:relighting}
\end{figure*}

\paragraph{Synthetic decomposed scene reconstruction.}
Since SceneSmith provides complete ground-truth meshes, its evaluation also covers hidden and contact surfaces.
Our method outperforms all baselines on every reconstruction metric and achieves the highest decomposition F1 at tight thresholds (Tab.~\ref{tab:scenesmith_objects}), indicating that joint $N$-volume generation recovers complete, accurately placed instances, including surfaces not visible in any input view.
At loose thresholds, Astra matches more objects, but its lower scores at tight thresholds show that it places them less precisely.

\paragraph{Textured reconstruction.}
Our texture stage generates PBR materials (base color, metallic, and roughness) for every object, so the decomposed scene can be relit and edited directly. 
Fig.~\ref{fig:relighting} shows qualitative decomposition, reconstruction, and relighting results on ScanNet++.


\subsection{Ablation Study}
\label{sec:ablation}

\paragraph{3D geometry condition.}
Tab.~\ref{tab:abl_geocond} ablates the 3D geometry condition of the sparse structure stage. 
For all four variants, we train the corresponding sparse structure model for 50k steps and use the same shape model.
A single TSDF shared across all volumes improves reconstruction accuracy but has little effect on decomposition, since it provides the same cues to every volume. 
Per-volume TSDFs break this symmetry and substantially improve decomposition at all thresholds. 
On real data, however, they degrade reconstruction, as errors in the predicted instance masks translate into contradictory per-volume cues. 
Training with mask augmentation makes the model robust to such errors: it yields the best decomposition and recovers most of the reconstruction quality.

\begin{table}[t]
  \centering
  \caption{\textbf{Ablation of the 3D geometric condition.} We evaluate the 3D reconstruction and decomposition metrics on 20 scenes from ScanNet++ given eight input images.}
  \label{tab:abl_geocond}
  \footnotesize
  \setlength{\tabcolsep}{5pt}
  \begin{tabular}{lcccccc}
    \toprule
    Geometric & CD $\downarrow$ & \multicolumn{5}{c}{Decomposition F1 $\uparrow$} \\
    \cmidrule(l){3-7}
    condition & (cm) & @1 & @2 & @4 & @8 & @16 \\
    \midrule
    None & \cellcolor{yellow!40}5.99 & \cellcolor{yellow!40}0.019 & 0.118 & 0.312 & 0.473 & 0.554 \\
    Shared TSDF & \cellcolor{red!30}\textbf{5.29} & 0.015 & \cellcolor{yellow!40}0.125 & \cellcolor{yellow!40}0.332 & \cellcolor{yellow!40}0.484 & \cellcolor{yellow!40}0.571 \\
    Per-Volume TSDF & 7.04 & \cellcolor{orange!35}0.030 & \cellcolor{orange!35}0.170 & \cellcolor{orange!35}0.395 & \cellcolor{orange!35}0.529 & \cellcolor{orange!35}0.617 \\
    + Mask aug. & \cellcolor{orange!35}5.85 & \cellcolor{red!30}\textbf{0.032} & \cellcolor{red!30}\textbf{0.174} & \cellcolor{red!30}\textbf{0.404} & \cellcolor{red!30}\textbf{0.545} & \cellcolor{red!30}\textbf{0.630} \\
    \bottomrule
  \end{tabular}
\end{table}

\paragraph{Number of volumes.}
The number of volumes $N$ trades separability against cost: global attention makes compute grow quadratically with $N$, while too few volumes force touching objects to share one.
Tab.~\ref{tab:abl_num_volumes} quantifies this on our training chunks.
With $N{=}2$, many objects cannot be separated, especially in cluttered scenes; $N{=}4$ separates almost all
objects, while $N{=}5$ gains little at a considerably higher cost, so we use $N{=}4$.
\begin{table}[t]
  \centering
  \caption{\textbf{Ablation on the number of volumes $N$.} Separable instances: share of objects that can still be extracted as separate instances. Nearby objects (within 1.3\% of the chunk size) that must share a volume merge into one instance. Cost: one denoiser forward pass.}
  \label{tab:abl_num_volumes}
  \small
  \setlength{\tabcolsep}{5pt}
  \begin{tabular}{@{}c ccc cc@{}}
    \toprule
    & \multicolumn{3}{c}{Separable instances (\%) $\uparrow$} & \multicolumn{2}{c}{Cost $\downarrow$} \\
    \cmidrule(lr){2-4} \cmidrule(l){5-6}
    $N$ & SAGE & InfiniGen & Physicl & Tokens & TFLOPs \\
    \midrule
    2 & 78.59 & 78.89 & 58.03 & 8{,}192 & 32.1 \\
    3 & 97.05 & 98.36 & 85.99 & 12{,}288 & 57.2 \\
    4 & 99.94 & 99.98 & 95.56 & 16{,}384 & 88.5 \\
    5 & 100.00 & 100.00 & 98.68 & 20{,}480 & 126.0 \\
    \bottomrule
  \end{tabular}
\end{table}
\subsection{Limitations}
Our method relies on multi-view consistent instance masks to assign objects to volumes. 
Mask augmentation makes it robust to moderate errors, but missed or merged masks still lead to missed or merged objects, and the mask predictor is the main source of errors on real data.
With only $N{=}4$ volumes, very cluttered scenes can force touching objects into a shared volume, where they merge
into one instance.
Finally, the voxel resolution of TRELLIS.2~\cite{xiang2025trellis2} at the chunk scale makes very small and thin
objects difficult to reconstruct accurately.

\section{Conclusion}
\label{sec:conclusion}
We presented InstanceRecon, which reconstructs indoor scenes from sparse posed images as complete PBR meshes of the individual objects and the room structure.
Since objects need to be separated only where they touch, we distribute them across a small, fixed number of volumes via proximity-weighted graph coloring and generate all volumes jointly with a single model built on TRELLIS.2, guided by a per-volume geometry condition.
Although trained only on synthetic data, our method outperforms all baselines, with the largest margins at tight thresholds, showing that joint generation places objects more precisely than compositional approaches.
We see this work as a step toward reconstructed scenes that are directly usable for editing, simulation, and content creation.

\section*{Acknowledgements}

Katharina Schmid is a doctoral fellow at the Munich Data Science Institute (MDSI), and Nicolas von Lützow was supported by the MDSI focus topic \textit{Understanding Existing Structures in Building Planning} (USP) as well as through the Bavarian AI Foundation Model Initiative, funded by the Bavarian State Ministry of Science and the Arts (StMWK).
Angela Dai was supported by the ERC Starting Grant SpatialSem (101076253).
This work was further supported by the ERC Consolidator Grant \textit{Gen3D} (101171131), by compute resources from the Jülich Supercomputing Center under project \textit{MeshFoundation}, and by the Computing Systems Lab, part of the Huawei Technologies Switzerland AG.

{
    \small
    \bibliographystyle{ieeenat_fullname}
    \bibliography{main}
}

\appendix
\clearpage
\setcounter{page}{1}
\maketitlesupplementary

\section{Additional Results}

\paragraph{Quantitative results including structure}
Tab.~\ref{tab:scannetpp_with_structure} and Tab.~\ref{tab:scenesmith_with_structure} show the reconstruction and decomposition results on ScanNet++ and SceneSmith for full scenes including structure (walls, floor, ceiling).
Since ShapeR, WorldSculpt, and ReplicateAnyScene don't output any structural components, they are not included in this comparison.

\begin{table*}[t]
  \centering
  \caption{\textbf{Quantitative evaluation on real-world data incl. structure.} We evaluate 3D reconstruction and decomposition metrics on 20 scenes from ScanNet++ given eight input images and predicted depth and instance masks.}
  \label{tab:scannetpp_with_structure}
  \small
  \setlength{\tabcolsep}{3pt}
  \begin{tabular}{lrrrrrrr@{\hspace{10pt}}rrrrr}
    \toprule
     & \multicolumn{7}{c}{Reconstruction} & \multicolumn{5}{c}{Decomposition} \\
    \cmidrule(lr){2-8} \cmidrule(lr){9-13}
     & \multirow{2}{*}{CD (cm) $\downarrow$} & \multicolumn{5}{c}{F-score $\uparrow$} & \multirow{2}{*}{NC $\uparrow$} & \multicolumn{5}{c}{F1 $\uparrow$} \\
    \cmidrule(lr){3-7} \cmidrule(lr){9-13}
    Method & & @1 & @2 & @4 & @8 & @16 & & @1 & @2 & @4 & @8 & @16 \\
    \midrule
    GenRecon+SpaCeFormer & \cellcolor{yellow!40}6.56 & \cellcolor{yellow!40}0.082 & \cellcolor{yellow!40}0.199 & \cellcolor{yellow!40}0.423 & \cellcolor{yellow!40}0.733 & \cellcolor{yellow!40}0.929 & \cellcolor{orange!35}0.845 & \cellcolor{yellow!40}0.001 & 0.017 & 0.107 & 0.286 & 0.436 \\
    GenRecon+MV3DIS & \cellcolor{yellow!40}6.56 & \cellcolor{yellow!40}0.082 & \cellcolor{yellow!40}0.199 & \cellcolor{yellow!40}0.423 & \cellcolor{yellow!40}0.733 & \cellcolor{yellow!40}0.929 & \cellcolor{orange!35}0.845 & \cellcolor{yellow!40}0.001 & \cellcolor{yellow!40}0.018 & \cellcolor{yellow!40}0.139 & \cellcolor{yellow!40}0.334 & \cellcolor{yellow!40}0.474 \\
    \midrule
    Astra & \cellcolor{orange!35}4.87 & \cellcolor{orange!35}0.145 & \cellcolor{orange!35}0.326 & \cellcolor{orange!35}0.597 & \cellcolor{orange!35}0.832 & \cellcolor{orange!35}0.960 & \cellcolor{yellow!40}0.827 & \cellcolor{orange!35}0.005 & \cellcolor{orange!35}0.053 & \cellcolor{orange!35}0.219 & \cellcolor{orange!35}0.480 & \cellcolor{red!30}\textbf{0.638} \\
    \midrule
    Ours & \cellcolor{red!30}\textbf{2.73} & \cellcolor{red!30}\textbf{0.272} & \cellcolor{red!30}\textbf{0.576} & \cellcolor{red!30}\textbf{0.831} & \cellcolor{red!30}\textbf{0.949} & \cellcolor{red!30}\textbf{0.985} & \cellcolor{red!30}\textbf{0.871} & \cellcolor{red!30}\textbf{0.027} & \cellcolor{red!30}\textbf{0.167} & \cellcolor{red!30}\textbf{0.412} & \cellcolor{red!30}\textbf{0.560} & \cellcolor{orange!35}0.631 \\
    \bottomrule
  \end{tabular}
\end{table*}

\begin{table*}[t]
  \centering
  \caption{\textbf{Quantitative evaluation on synthetic data incl. structure.} We evaluate 3D reconstruction and decomposition metrics on 20 scenes from SceneSmith given eight input images and predicted depth and instance masks.}
  \label{tab:scenesmith_with_structure}
  \small
  \setlength{\tabcolsep}{3pt}
  \begin{tabular}{lrrrrrrr@{\hspace{10pt}}rrrrr}
    \toprule
     & \multicolumn{7}{c}{Reconstruction} & \multicolumn{5}{c}{Decomposition} \\
    \cmidrule(lr){2-8} \cmidrule(lr){9-13}
     & \multirow{2}{*}{CD (cm) $\downarrow$} & \multicolumn{5}{c}{F-score $\uparrow$} & \multirow{2}{*}{NC $\uparrow$} & \multicolumn{5}{c}{F1 $\uparrow$} \\
    \cmidrule(lr){3-7} \cmidrule(lr){9-13}
    Method & & @1 & @2 & @4 & @8 & @16 & & @1 & @2 & @4 & @8 & @16 \\
    \midrule
    GenRecon+SpaCeFormer & \cellcolor{yellow!40}9.13 & 0.078 & 0.164 & 0.317 & 0.585 & \cellcolor{yellow!40}0.833 & \cellcolor{yellow!40}0.812 & 0.003 & 0.057 & 0.197 & 0.336 & 0.469 \\
    GenRecon+MV3DIS & 9.18 & \cellcolor{yellow!40}0.080 & \cellcolor{yellow!40}0.167 & \cellcolor{yellow!40}0.327 & \cellcolor{yellow!40}0.599 & \cellcolor{orange!35}0.847 & 0.804 & \cellcolor{yellow!40}0.009 & \cellcolor{yellow!40}0.083 & \cellcolor{yellow!40}0.227 & \cellcolor{yellow!40}0.379 & \cellcolor{yellow!40}0.532 \\
    \midrule
    Astra & \cellcolor{orange!35}3.49 & \cellcolor{orange!35}0.277 & \cellcolor{orange!35}0.430 & \cellcolor{orange!35}0.646 & \cellcolor{orange!35}0.926 & \cellcolor{red!30}\textbf{0.982} & \cellcolor{orange!35}0.858 & \cellcolor{orange!35}0.073 & \cellcolor{orange!35}0.298 & \cellcolor{orange!35}0.610 & \cellcolor{red!30}\textbf{0.776} & \cellcolor{red!30}\textbf{0.842} \\
    \midrule
    Ours & \cellcolor{red!30}\textbf{2.34} & \cellcolor{red!30}\textbf{0.414} & \cellcolor{red!30}\textbf{0.703} & \cellcolor{red!30}\textbf{0.880} & \cellcolor{red!30}\textbf{0.953} & \cellcolor{red!30}\textbf{0.982} & \cellcolor{red!30}\textbf{0.888} & \cellcolor{red!30}\textbf{0.226} & \cellcolor{red!30}\textbf{0.497} & \cellcolor{red!30}\textbf{0.671} & \cellcolor{orange!35}0.751 & \cellcolor{orange!35}0.786 \\
    \bottomrule
  \end{tabular}
\end{table*}

\begin{figure*}[t]
  \centering
  \setlength{\tabcolsep}{1pt}
  \newcommand{\litw}{\dimexpr(\textwidth-8\tabcolsep)/5\relax}
  \newcommand{\litrow}[1]{%
    \includegraphics[width=\litw]{figures/lit/#1_input.jpg} &
    \includegraphics[width=\litw]{figures/lit/#1_decomp.jpg} &
    \includegraphics[width=\litw]{figures/lit/#1_normal.jpg} &
    \includegraphics[width=\litw]{figures/lit/#1_pbr_light1.jpg} &
    \includegraphics[width=\litw]{figures/lit/#1_pbr_light2.jpg} \\
  }
  \renewcommand{\arraystretch}{0.5}
  \begin{tabular}{@{}ccccc@{}}
    \small Input image & \small Decomposition & \small Normals & \small Lit v1 & \small Lit v2 \\[3pt]
    \litrow{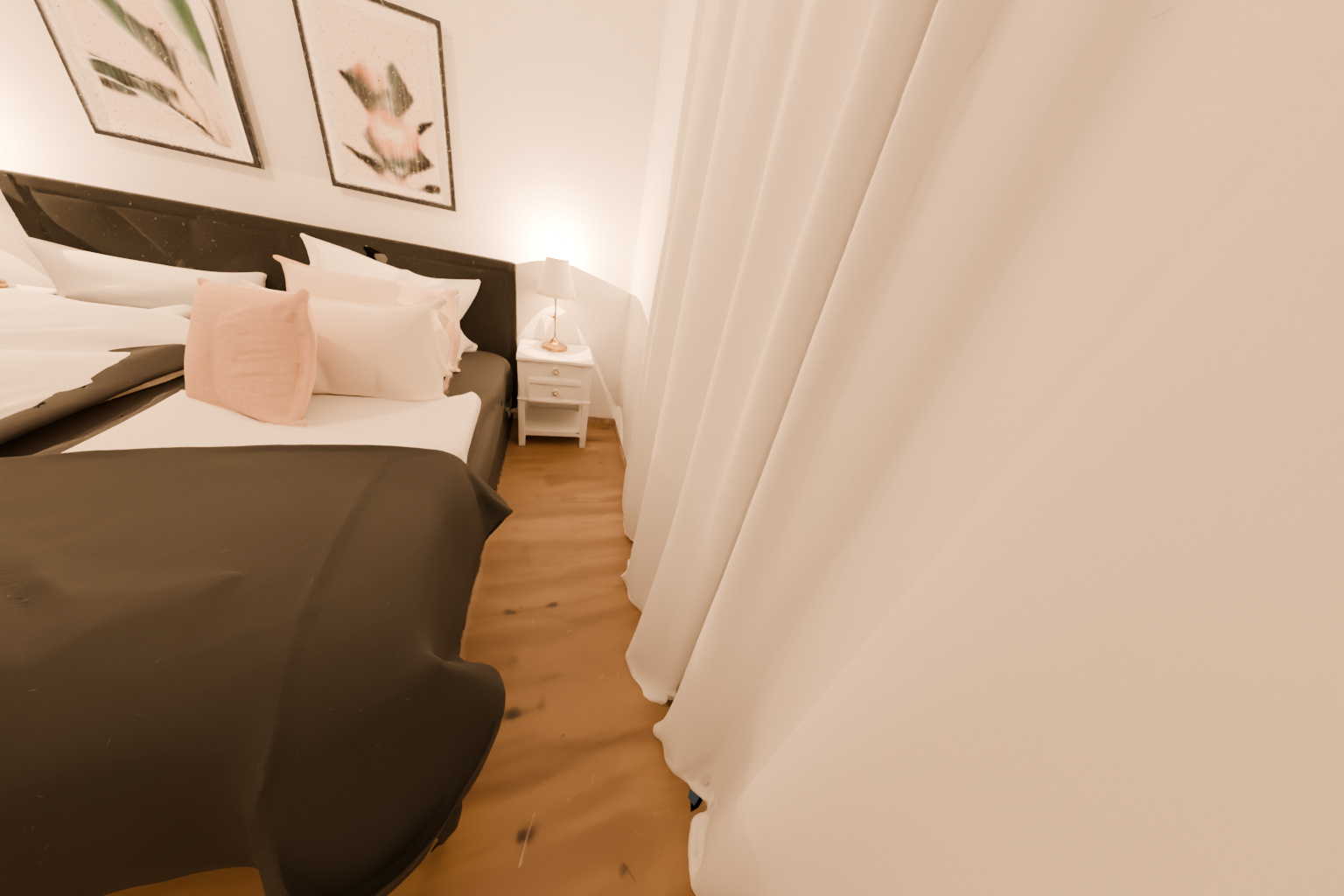}
    \litrow{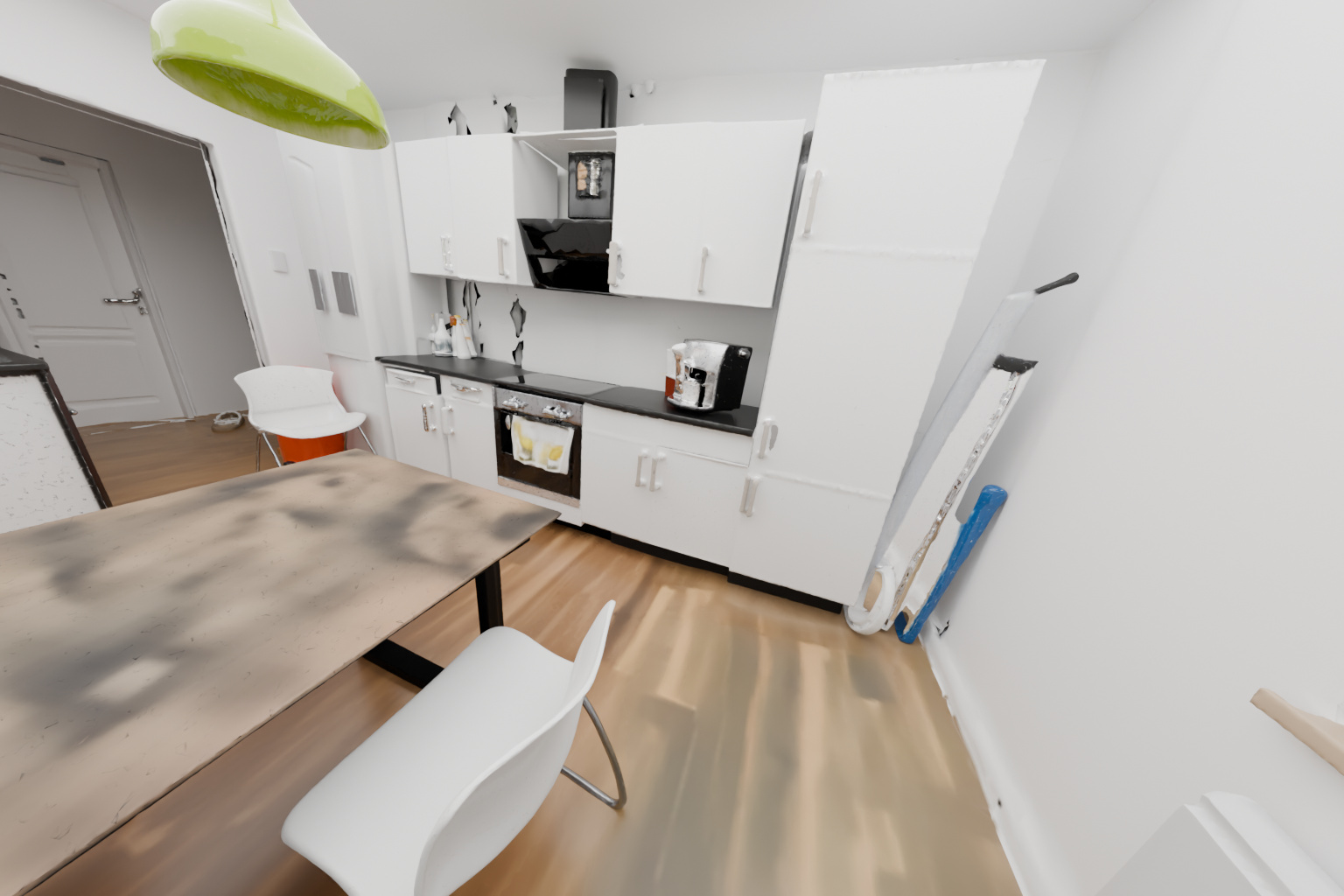}
    \litrow{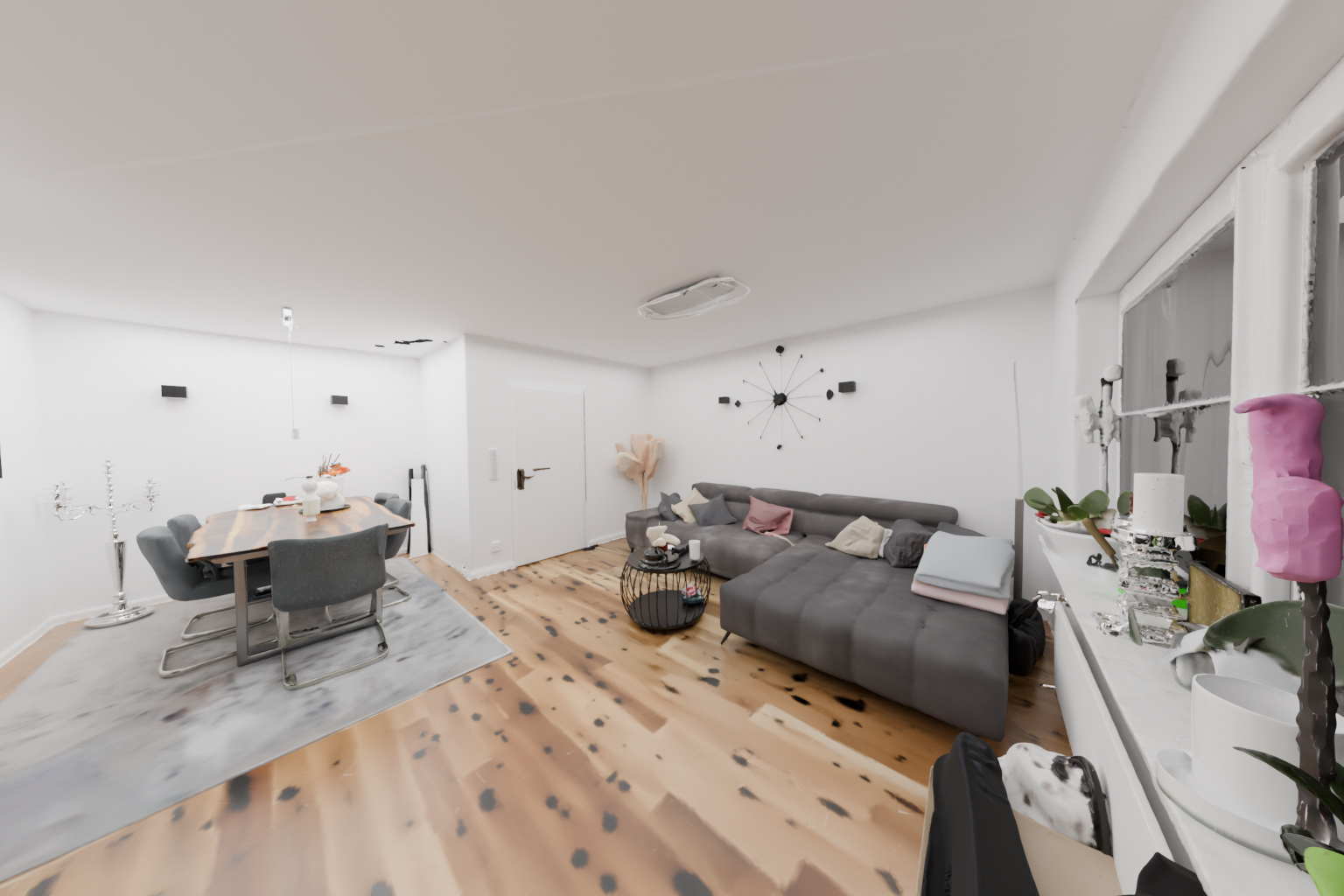}
    \litrow{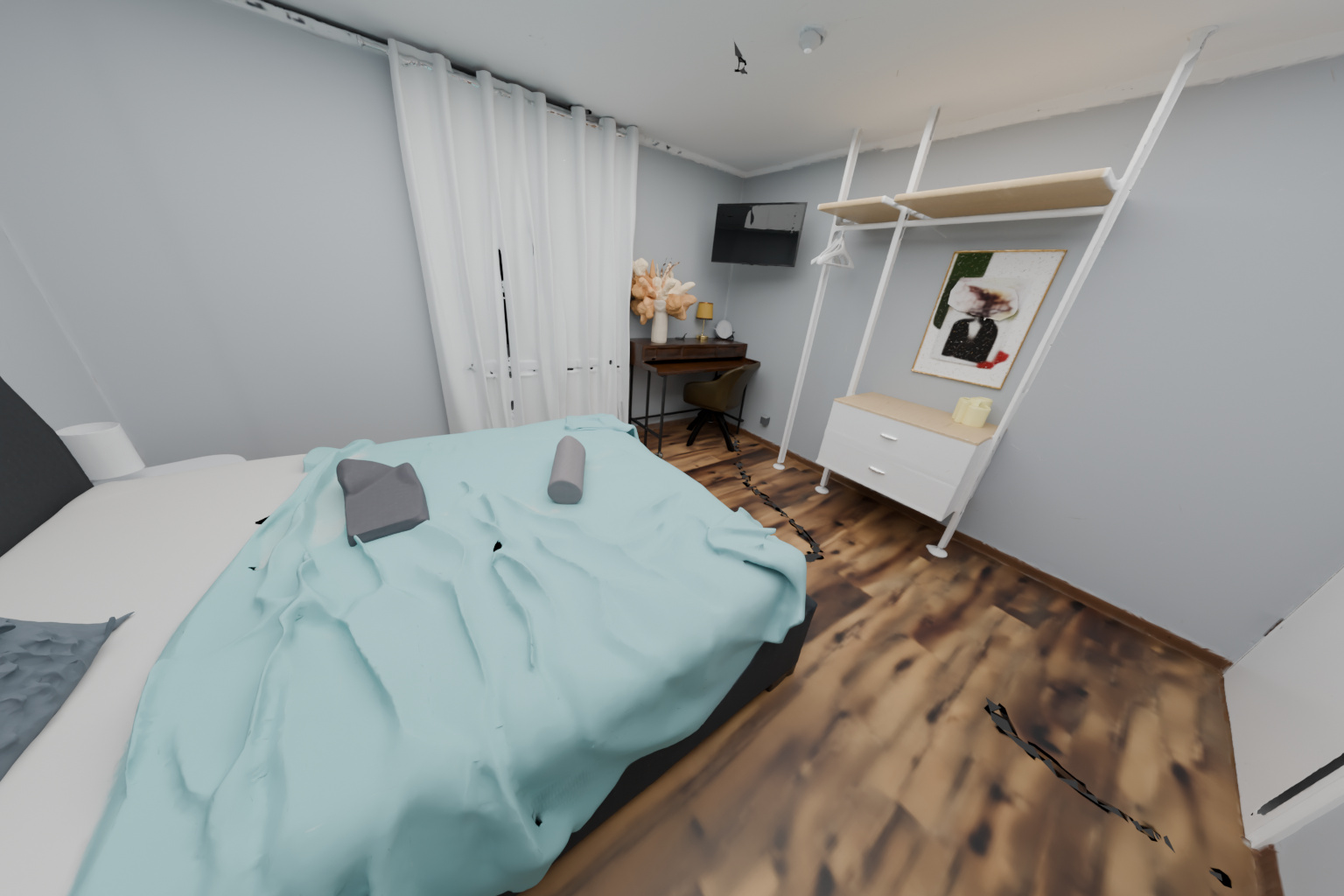}
    \litrow{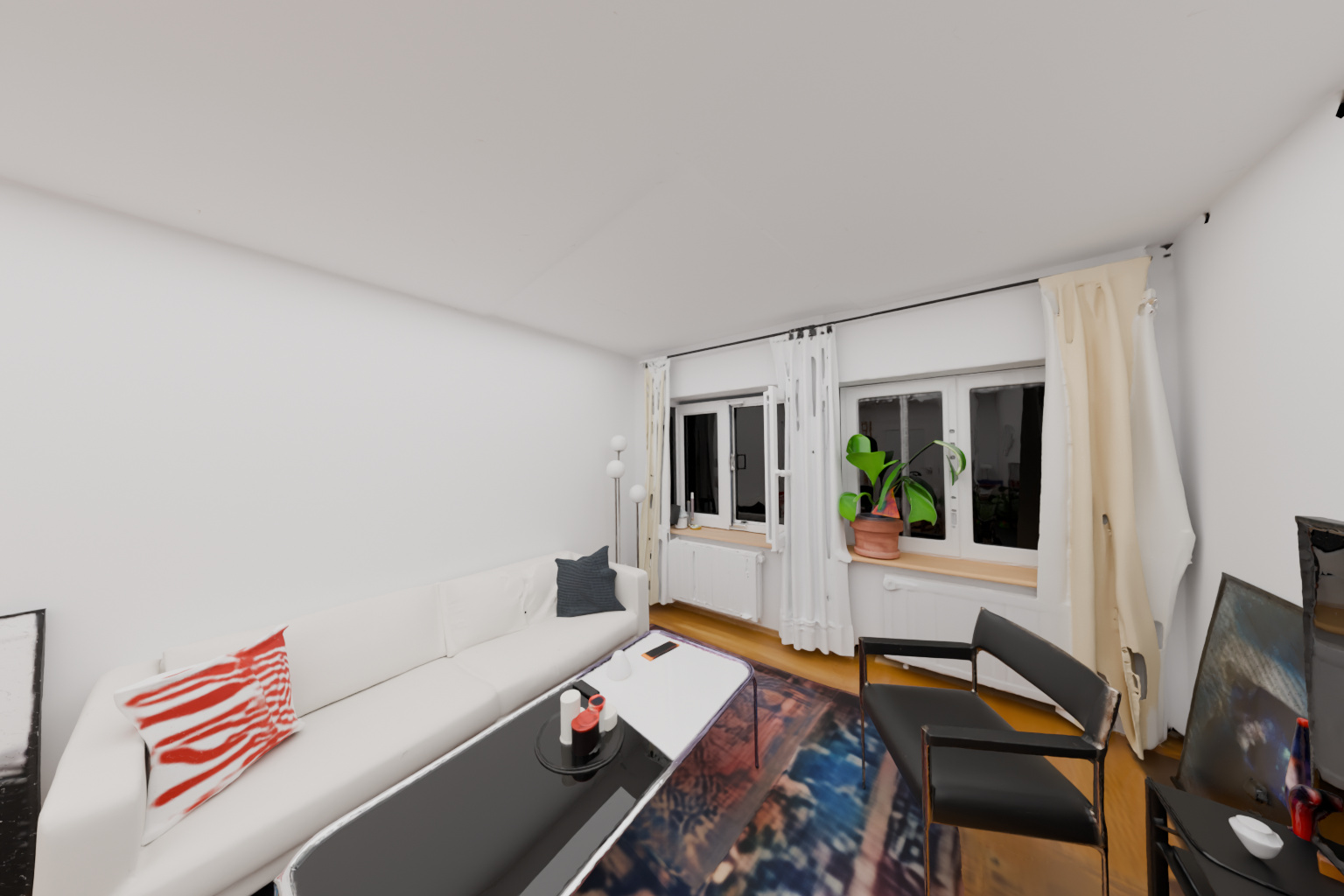}
  \end{tabular}
  \caption{\textbf{Qualitative decomposition, reconstruction, and relighting results on real-world data.} Results on ScanNet++ from eight input images and predicted depth and instance masks. Our approach reconstructs highly realistic relightable textures.}
  \label{fig:relighting2}
\end{figure*}

\paragraph{Additional qualitative results}
Fig.~\ref{fig:relighting2} provides additional qualitative results on the decomposition, reconstruction and texturing quality.

\section{Experimental Setup}

\paragraph{Data preprocessing.}
For ScanNet++~\cite{yeshwanthliu2023scannetpp}, we take eight DSLR frames evenly spaced among the frames not flagged as bad, at $1536{\times}1024$\,px, with the official poses in the mesh-aligned frame. 
For SceneSmith~\cite{scenesmith2026}, we render eight $1024{\times}1024$\,px views with Cycles from cameras placed randomly inside the room.
All methods receive the same views and ground-truth poses.
Metric depth is predicted with Depth Anything~3 (DA3-Nested-Giant-Large-1.1)~\cite{depthanything3} in one posed pass over all views at \SI{1008}{px}.
For instance masks, we run SAM3~\cite{carion2026sam3segmentconcepts} on every view, prompted with the scene's ground-truth category names plus \emph{floor}, \emph{ceiling} and \emph{wall}, keep masks with a score above $0.4$ and at least \SI{100}{px}, and remove detections that lie mostly inside a
higher-scoring one.
Each mask is lifted to 3D with the predicted depth, excluding pixels at depth discontinuities.
Two detections from different views match if at least half of their points, reprojected into the other view, land inside the other mask within $\max(\SI{3}{\centi\meter}, 3\%)$ of the observed depth; occluded points are ignored.
Matches are grouped greedily, strongest first, with at most one mask per view in each group, and each group takes the
category with the highest total score.
Groups whose 3D boxes overlap (IoU${}>0.4$) without sharing a view are merged, and single-view instances with a score below $0.6$ are discarded. Floor, ceiling, and wall each receive a fixed ID.
The ID maps are downsampled to the \SI{256}{px} model input by nearest-neighbor sampling.

\paragraph{Conditioning networks.}
For the image condition, DINOv3 ViT-L/16~\cite{siméoni2025dinov3} patch tokens at \SI{1024}{px} are projected onto a grid twice as fine as the latent grid ($32^3$ for the sparse structure stage, $64^3$ for shape and texture).
In each voxel, the features of the observing views are aggregated with IBRNet-style~\cite{wang2021ibrnet} weights: an MLP scores each view's features along with the mean and variance across views, followed by a softmax over the views.
The result is independent of the order and number of views.
Two coverage channels (observed by any view, fraction of observing views) are appended, and residual 3D convolution blocks with a space-to-channel step reduce the grid to the latent resolution.
For the geometry condition of the sparse structure stage, we fuse one projective TSDF per volume and one for unassigned pixels at $128^3$ per chunk (truncation of 4 voxels, plus $\log(1{+}n)$ of the observation count $n$).
Depth pixels at depth jumps (${>}5\%$ within $3{\times}3$) are excluded, and one-pixel mask borders are treated as unassigned.
Each volume's TSDF, concatenated with the unassigned TSDF, is encoded by a network shared across volumes: a lossless space-to-channel fold to $64^3$, and two levels of residual 3D convolution blocks, each followed by a space-to-channel step (widths 64, 128, 512) down to $16^3$. 
Both conditions are injected before every transformer block through zero-initialized linear layers.

\paragraph{Training.}
We use $N{=}4$ volumes, with volume embeddings initialized as the vertices of a regular simplex with norm 2, randomly permute volumes 1 to 3 in every sample, and condition each chunk on up to 16 views.
All stages use AdamW ($\beta{=}(0.9, 0.95)$, weight decay $0.01$ on new modules)~\cite{loshchilov2019decoupledweightdecayregularization}, linear warm-up over 3k steps, an EMA rate of $0.9995$, bfloat16 mixed precision, and adaptive gradient clipping.
The sparse structure stage samples timesteps from a logit-normal distribution (mean 1, std 1) and, per sample, drops both conditions with probability $0.05$ and only the image or only the geometry condition with probability $0.05$ each.
The shape and texture stages sample timesteps uniformly and drop the condition with probability $0.1$.
To make the sparse structure stage robust to predicted masks, we corrupt the ground-truth masks with the error types of predicted masks (Tab.~\ref{tab:mask_aug}); floor, ceiling, and walls are never removed, relabeled, or split themselves, but their pixels can be covered by dilated objects and spurious masks.

\begin{table}[t]
  \centering
  \footnotesize
  \setlength{\tabcolsep}{3pt}
  \caption{\textbf{Mask augmentation.} Error types applied to the ground-truth instance masks during training, in
    this order. Probabilities are per object and view unless noted; areas are relative to the view.}
  \label{tab:mask_aug}
  \begin{tabularx}{\linewidth}{@{}l X l@{}}
    \toprule
    Error type & Effect & Prob. \\
    \midrule
    Missed instance & removed in all views & 0.05--0.45\textsuperscript{*} \\
    Missed in view & removed in one view & 0.05 \\
    Absorbed & small object (${<}2\%$) takes the ID of a larger touching object & 0.25 \\
    Boundary jitter & erode / keep / dilate by one pixel & .15 / .55 / .30 \\
    Wrong volume & assigned to another volume & 0.03 \\
    Split & a straight cut moves 20--45\% of the mask to a new instance in another volume & 0.05 \\
    Spurious mask & ellipse (0.03--3\%) with a random volume on non-object pixels & Pois(0.2)\textsuperscript{**} \\
    Unlabeled view & all pixels of the view unassigned & 0.02\textsuperscript{**} \\
    \bottomrule
  \end{tabularx}
  \par\smallskip
  {\raggedright\scriptsize
   \textsuperscript{*}\,Per object: 0.45 / 0.2 / 0.1 / 0.05 for a mean visible area ${<}0.1\%$ / ${<}0.5\%$ / ${<}2\%$ /
   larger. \textsuperscript{**}\,Per view.\par}
\end{table}

\paragraph{Inference.}
We tile the scene with a single horizontal layer of cubic chunks.
To size them, we lift all depth pixels to 3D and take the floor and ceiling heights as the 0.2th and 99.97th percentiles of the point heights; the chunk edge is 1.05 times their difference, so that one chunk spans the room from floor to ceiling.
The horizontal region to tile is given by the 0.5th and 99.5th percentiles of the points along each horizontal axis.
Chunk positions with fewer than 5\% of the median number of depth points per occupied position are dropped, and random shifts are multiples of one latent voxel.
For the volume assignment, instances are lifted into voxels of $1/256$ chunk edge from their 1-px eroded masks and dilated by 8 voxels, so instances within $1/16$ chunk edge (about \SI{18}{\centi\meter}) are in contact; CP-SAT~\cite{cpsat} solves the coloring with a \SI{10}{\second} time limit. Tab.~\ref{tab:sampling} lists the sampling parameters. 
After decoding, hole filling is applied.
For instance extraction, surfaces are voxelized at $0.0065$ chunk edges, and components below $0.003$~chunk~edges~$^2$ are merged into a large component within 2 voxels or discarded.
Two instances from different volumes that each cover at least 80\% of the other's surface within $0.01$ chunk edges are duplicates, and the smaller one is removed.
PBR textures are baked per instance.

\begin{table}[t]
  \centering
  \small
  \caption{\textbf{Sampling parameters.} All stages use 12 steps and apply guidance for $t \in [0.6, 1]$. For the sparse structure stage, the
    guidance scale is given as $s_\text{img}$\,/\,$s_\text{geo}$.}
  \label{tab:sampling}
  \newcommand{\hd}[1]{\begin{tabular}[b]{@{}c@{}}#1\end{tabular}}
  \begin{tabular}{@{}lccc@{}}
    \toprule
    Stage & \hd{Guidance\\scale} & \hd{Guidance\\rescale} & \hd{Time\\shift} \\
    \midrule
    Sparse structure & 3\,/\,7.5 & 0.7 & 5 \\
    Shape & 7.5 & 0.5 & 1.5 \\
    Texture & 3 & 0 & 3 \\
    \bottomrule
  \end{tabular}
\end{table}


\section{Evaluation Details}

\paragraph{Common preprocessing.}
Every method must output one mesh per instance in the world frame of the dataset; no alignment is performed, so errors
in pose and scale are penalized.
All methods, including ours, pass through the same steps before scoring.
First, predicted faces whose centroid lies outside the GT room are removed.
The room is the bounding box of the GT walls, floor, and ceiling, rotated to the wall orientation and enlarged by \SI{30}{\centi\meter}.
This step is needed because some methods (\eg Astra) also reconstruct buildings seen through windows, which would otherwise dominate the accuracy terms.
Second, instances supported by fewer than $5$ points are dropped and count neither as false positives nor as false negatives.

\paragraph{Surface protocol (SceneSmith).}
SceneSmith~\cite{scenesmith2026} provides complete GT geometry, so we compare full surfaces.
Both prediction and GT are sampled area-uniformly with a fixed seed at $10$k points per \si{\square\meter}; each instance receives at least $10$k points in the decomposition.
Distances are measured point-to-point, from each sample to its nearest neighbor on the other side.
SceneSmith models walls, floor, and ceiling as thick slabs. 
Their backs, tops, and end caps are never visible from inside the room, so for both GT and prediction we only score the inner
shell.
It consists of faces that have \SI{25}{\centi\meter} of free space on their room-facing side, lie within the room's height range (\SI{\pm 10}{\centi\meter}) and above the floor, and are not enclosed by another slab.

\paragraph{Ray protocol (ScanNet++).}
The ScanNet++~\cite{yeshwanthliu2023scannetpp} GT mesh is incomplete, so we only score surfaces observed by the laser scanner. 
We treat its scan positions as ray origins and its returns as GT points.
A return is dropped if the first GT-mesh surface along its ray lies more than \SI{5}{\centi\meter} in front of it, as is the case for mirrors, glass, and moved doors.
It is also dropped if it lies on an anonymized or removed region.
The remaining returns take the instance label and normal of the first GT-mesh hit within \SI{2}{\centi\meter}; returns without such a hit, or whose hit has no instance label, are \emph{void} and are ignored in all object scores.
To make every observed patch count exactly once, returns are pooled per scan position into one ray per cell (edge length $h/128$ for room height $h$) and instance, aimed at the mean return.
If several scan positions observe the same cell, only the most frontal ray (largest cosine to the surface normal) is kept.
Each remaining ray is cast into the prediction, and its first hit becomes the predicted point; a ray without a hit contributes no predicted point.
In the object-only variant, rays are cast into the prediction with its walls, floor, and ceiling removed, so predicted structure cannot hide a misplaced object.
Rays whose GT return lies on the structure do not provide GT points but still score the predicted objects they hit.
As a consequence, predicted geometry is only scored where a ray hits it first; geometry that no ray reaches is not penalized.

\paragraph{Metrics.}
For reconstruction, let $d_\text{acc}$ be the distance from each predicted point to the nearest GT point and $d_\text{comp}$ the distance from each GT point to the nearest predicted point.
Accuracy and completeness are their means, and the Chamfer distance is $\text{CD} = (\text{acc} + \text{comp})/2$.
Precision and recall at threshold $\tau$ are the fractions of $d_\text{acc}$ and $d_\text{comp}$ below $\tau$; F@$\tau$ is their harmonic mean, with $\tau \in \{1, 2, 4, 8, 16\}$\,\si{\centi\meter}.
Normal consistency (NC) is the mean absolute cosine between the normals of corresponding points.
For decomposition, each instance is represented by its point set (surface samples or ray points).
The cost of pairing a GT instance with a predicted instance is their two-sided Chamfer distance.
For each $\tau$, we compute a one-to-one Hungarian matching that maximizes $\sum(\tau - \text{CD})$ over pairs with $\text{CD} < \tau$. 
Matched pairs are true positives, unmatched predictions are false positives, and unmatched GT instances are false negatives, which yields precision, recall, and F1@$\tau$ for $\tau \in \{1, 2, 4, 8, 16\}$\,\si{\centi\meter}.

\paragraph{Aggregation and reporting.}
All metrics are computed per scene and averaged over scenes, so every scene carries equal weight regardless of its size.
Our main results score objects only: GT and predicted walls, floor, and ceiling are removed.
In the ray protocol, rays whose GT return lies on the structure provide no GT points but are still cast into the prediction, so a predicted object in front of a wall counts against accuracy and as a false positive.
We additionally report the full scene, in which walls, floor, and ceiling are merged into a single \emph{structure} instance on both sides; this variant applies only to methods that predict structure.
Predicted surfaces that belong to no instance (SpaCeFormer, MV3DIS) count only towards the full-scene reconstruction and are excluded from all other scores.
Since the two protocols compare different point sets (densely sampled surfaces vs.\ scanned rays), numbers are not comparable across datasets.


\section{Baseline Implementation Details}

All methods receive the same input per scene: the 8 RGB views with the dataset's camera poses and intrinsics.
ShapeR~\cite{siddiquiShapeRRobustConditional2026}, ReplicateAnyScene~\cite{dongReplicateAnySceneZeroShotVideoto3D2026}, WorldSculpt~\cite{niuWorldSculptGeneratingCompositional2026}, and Astra~\cite{openai2026gpt6astra} receive the same predicted, multi-view-consistent instance masks and the same Depth Anything~3 depth maps, predicted jointly for all views with the given cameras. 
The two decomposition methods (MV3DIS~\cite{zhao2026mv3dismultiviewmaskmatching} and SpaCeFormer~\cite{choy2026spaceformer}) operate on the GenRecon~\cite{schmid2026genreconbridginggenerativepriors} mesh.
No ground-truth geometry is used.
All outputs are evaluated in the dataset's metric world frame without any alignment of position, rotation, or scale.
ShapeR, ReplicateAnyScene and WorldSculpt reconstruct objects only, whereas GenRecon (and its decompositions) and Astra also reconstruct the room structure.

\paragraph{GenRecon.}
We run the official implementation of GenRecon~\cite{schmid2026genreconbridginggenerativepriors} with the publicly released checkpoints and without code modifications, using its default inference settings. 
GenRecon receives the same 8 posed views as all other methods, with the dataset's camera poses and intrinsics, through the posed-image (smartphone) input mode of the official code, which splits each non-square view into two square crops.
The original method places its chunks using a sparse COLMAP point cloud of the dense input capture. 
Since only 8 views are available in our setting, we instead back-project the Depth Anything~3~\cite{depthanything3} depth maps of the input views into the world frame and downsample them on a \SI{1}{\centi\meter} voxel grid; these points are used only to determine the
chunk layout. 
As output, we use GenRecon's merged scene mesh. 
Since GenRecon does not decompose the scene, its result is a single mesh that contains both objects and structural elements.

\paragraph{MV3DIS.}
To decompose the GenRecon reconstruction by lifting 2D instance masks, we apply MV3DIS~\cite{zhao2026mv3dismultiviewmaskmatching} to the GenRecon~\cite{schmid2026genreconbridginggenerativepriors} mesh, using the official code with
its default hyperparameters. 
MV3DIS expects a dense posed RGB-D sequence together with a scan mesh, whereas our 8 input views cover only part of the reconstructed geometry.
We therefore render 128 RGB-D views of the GenRecon mesh from virtual cameras inside the room (input image resolution, \ang{90} horizontal field of view), selected greedily from candidates on a \SI{0.5}{\meter} grid at three heights and six viewing directions so as to maximize the number of points seen in at least three views; all rendered views are used (instead of every 10th frame).
Since GenRecon's colors are albedo, the images are rendered with simple headlight shading so that adjacent surfaces of the same color remain distinguishable.
The 3D points are the mesh vertices downsampled on a 2\,cm voxel grid, and the superpoints are computed with the ScanNet mesh segmentator using ScanNet's parameters. 
The 2D masks come from the official GroundingDINO~\cite{liu2024groundingdinomarryingdino} + SAM2.1~\cite{ravi2024sam2segmentimages} mask generator with its default settings and the ScanNet200 vocabulary~\cite{dai2017scannet}, extended by \emph{wall} and \emph{floor} so that these surfaces are segmented as well.
Since MV3DIS outputs class-agnostic regions that also cover structural elements, we assign each region the label with the most votes from the 2D masks, discard regions with fewer than 20 points, regions not visible in any view, and regions of which less than \SI{40}{\percent} of the visible observations are covered by a 2D mask, and treat regions labeled as wall, floor, ceiling or one of their variants as structure. 
Each face is assigned to the instance shared by at least two of its vertices; the remaining faces are left unassigned and ignored in the evaluation.
The number of views and the field of view were chosen on a single ScanNet++~\cite{yeshwanthliu2023scannetpp} scene.

\paragraph{SpaCeFormer.}
To decompose the GenRecon reconstruction with a 3D instance segmenter, we apply SpaCeFormer~\cite{choy2026spaceformer} to the GenRecon~\cite{schmid2026genreconbridginggenerativepriors}  mesh, using the official code and released checkpoint without modifications, including the default post-processing (mask NMS, minimum of 20 points per mask). 
As input, we use the GenRecon mesh vertices with their base color, downsampled on a \SI{2}{\centi\meter} voxel grid (the model's own voxel size). 
The released pipeline outputs a ranked set of overlapping masks intended for AP evaluation.
To obtain a decomposition, we keep all masks with score (objectness $\times$ mask quality) $\geq 0.3$, assign each point to the highest-scoring mask containing it, and discard instances with fewer than 20
points. 
Masks whose top-1 open-vocabulary label among the ScanNet200~\cite{dai2017scannet} classes is a structural class (wall, floor, ceiling and their variants) take part in this assignment but are then treated as structure. 
Object labels are then propagated to unassigned points over 5 rounds of neighbor majority voting (3 cm radius); points in structural masks are not changed.
Finally, the point labels are transferred back to the mesh, and each face is assigned to the instance shared by at least two of its vertices; the remaining faces are left unassigned and ignored in the evaluation. 
The score threshold and hole filling were chosen on a single ScanNet++~\cite{yeshwanthliu2023scannetpp} scene that is not part of the evaluation set.

\paragraph{ShapeR.}
ShapeR~\cite{siddiquiShapeRRobustConditional2026} reconstructs each object separately from a sparse metric point cloud, a few posed grayscale views with the projections of the visible SLAM points, and a text caption. 
We use the official code and checkpoints with the \emph{quality} preset (16 views, 50 denoising steps) and a fixed seed, and provide each object in the input format of the authors' Depth Anything~3 variant.
The object's points are obtained by lifting its predicted instance mask in every view with the DA3~\cite{depthanything3} depth, expressed in a frame centered at its bounding box, filtered with the authors' DBSCAN step, and subsampled to 1024 points by farthest-point sampling.
In place of the SLAM point projections, each view marks the final points visible in it (inside the
instance mask and consistent with the DA3 depth).
The caption is the predicted category of the instance.
Since the released variant supports only a single image, we extend its loader to use all views of an object: views with fewer than 15 visible points are discarded, the rest are ranked by this number, and they are repeated cyclically to fill the 16 views.
Objects with fewer than 16 points after DBSCAN or without any remaining view are skipped (3\% of all objects, mostly tiny ones such as sockets and switches).
The resulting meshes are transformed back to the world frame.

\paragraph{ReplicateAnyScene.}
ReplicateAnyScene~\cite{dongReplicateAnySceneZeroShotVideoto3D2026} builds a compositional scene from a video: it estimates cameras and depth with VGGT~\cite{wang2025vggt}, discovers and tracks objects with a VLM and SAM3~\cite{carion2026sam3segmentconcepts}, generates each instance with SAM3D~\cite{teamSAM3D3Dfy2025} from its best view, aligns it to the observations (visual-spatial alignment), and refines the poses using physical relations such as floor support and wall attachment.
The public code is incomplete: object discovery is replaced by a user-provided category list, and the visual-spatial alignment is missing. 
Since our input consists of 8 posed views rather than a video, we replace the stages before generation with our inputs: the dataset cameras and the metric DA3~\cite{depthanything3} depth instead of VGGT, and the predicted instance masks, which are consistent across views, instead of SAM3 tracking and deduplication.
For the room frame needed for the refinement, we use the dataset's up axis, the floor height from a plane fitted to the lifted floor mask, and the orientation of planes fitted to the lifted wall mask.
In place of the VLM, each object's relation (supported by floor, attached to wall, embedded in wall, or supported by another object) is determined by simple geometric rules on its lifted points and its category.
Best-view selection, SAM3D generation, and the relation-based refinement are run unchanged.
We implement the visual-spatial alignment following the paper: starting from the SAM3D pose, we render the object into all views in which its instance mask appears, compute reciprocal MASt3R~\cite{mast3r_eccv24} matches between the renderings and the input images within the instance mask, lift them to 3D with the rendered and the DA3 depth, and update the pose with a similarity transform (Umeyama).
After 5 iterations, we keep the pose with the highest mean mask IoU.
Since all inputs are metric, no global scale alignment is needed; each object's scale is predicted by SAM3D and adjusted during alignment.

\paragraph{WorldSculpt.}
WorldSculpt~\cite{niuWorldSculptGeneratingCompositional2026} reconstructs each object from posed RGB images, instance masks, and a 3D bounding box, and composes the objects into a scene.
We run the official code and released checkpoints unchanged, with the settings of the official instructions (geometry only, without texturing).
As input, we provide the 8 posed views, the predicted instance masks, and, for each object, a gravity-aligned oriented bounding box fitted to its points lifted from the masks using the DA3~\cite{depthanything3} depth; no ground-truth geometry is used. 
WorldSculpt uses a view of an object only if its mask covers at least $\max(500\,\text{px}, \SI{0.1}{\percent})$ of the image, and skips objects without any such view.
With only 8 views, this excludes about \SI{20}{\percent} of all objects, mostly small ones, which are therefore missing from its reconstructions.

\paragraph{Astra.}
We run GPT-6-Astra~\cite{openai2026gpt6astra} with high reasoning effort in the Codex CLI~\cite{openai2026gpt6astra}. 
It builds the scene in a live Blender~5.2 session through the Blender MCP server~\cite{ahuja2025blendermcp}, restricted to scene inspection, screenshots, and Python execution, plus a sandboxed shell.
The Blender scene starts with one camera per view at its exact pose and intrinsics. 
Each scene is a single session without human interaction, with a 3\,h limit and web search disabled. 
Each session is isolated, so the agent sees only its workspace. 
We use the same prompt for all scenes (\cref{lst:astra_prompt}). 
For evaluation, each top-level object of the saved scene becomes one instance mesh, and the contents of the structure collection become the structure.
A run took 20\,min (median).

\lstset{basicstyle=\ttfamily\footnotesize, breaklines=true, columns=fullflexible, frame=single, upquote=true}
\begin{lstlisting}[caption={Astra prompt.}, label={lst:astra_prompt}]
Reconstruct the room in the photos as a 3D scene in Blender, using the Blender MCP tools.

The open Blender scene has one camera per photo, at its real pose, with the photo as its background image. The photos are in `images/`; poses and intrinsics are in `cameras.json` (metres, z up). For every photo there is an instance mask in `masks/` and a depth map in `depth/`; `masks/README.md` and `depth/README.md` describe their format.

Model the walls, floor, ceiling, doors, windows and every visible object, at real scale, so the scene matches the photos in shape and appearance from these cameras and from new viewpoints. Give each object a descriptive name and a matching material. When an object is made of several parts, parent all its parts to one empty named after the object. Give each physical object exactly one top-level object (an empty or a mesh). Put the walls, floor and ceiling in a collection named Structure. Do not change the cameras. Work only in the current directory.

Use the Cycles render engine, and set up the lights, world and render settings for it in scene.blend.

Save the scene as `scene.blend` in the current directory. When you stop, the saved `scene.blend` is your final result.
\end{lstlisting}


\end{document}